%% file: main.tex
\documentclass[journal ]{new-aiaa}
\usepackage[utf8]{inputenc}
\usepackage{textcomp}
\usepackage{graphicx}
\usepackage{amsmath}
\usepackage[version=4]{mhchem}
\usepackage{siunitx}
\usepackage{longtable,tabularx}
\usepackage{float}

\newtheorem{lemma}{Lemma}

\newtheorem{definition}{Definition}

\newenvironment{proof}[1][Proof]{\par\noindent\textbf{#1. }\ignorespaces}{\hfill$\square$\par}

\usepackage{subfiles}
\usepackage[ruled,lined,linesnumbered]{algorithm2e}
\usepackage[caption=false,font=normalsize,labelfont=sf,textfont=sf]{subfig}
\usepackage{textcomp}
\usepackage{url}
\usepackage{verbatim}
\usepackage{xcolor}
\usepackage{soul}

\sethlcolor{yellow}

\soulregister\cite7
\soulregister\ref7
\soulregister\eqref7
\soulregister\textbf7
\soulregister\textit7
\soulregister\emph7

\renewcommand{\hl}[1]{#1}

\usepackage{graphicx}
\usepackage{balance}
\usepackage{booktabs}
\usepackage{pifont}
\usepackage{adjustbox}

\newcommand{\cmark}{\ding{51}} % checkmark
\newcommand{\xmark}{\ding{55}} % crossmark

\title{Prognostics for Autonomous Deep-Space Habitat Health Management under Multiple Unknown Failure Modes}
\author{
Benjamin Peters\footnote{Assistant Professor, Department of Manufacturing and Industrial Engineering.}
}
\affil{University of Texas at Rio Grande Valley, Edinburg, TX 78539, USA}

\author{
Ayush Mohanty\footnote{Ph.D. Candidate, H. Milton Stewart School of Industrial and Systems Engineering; Corresponding Author, \href{mailto:ayush.mohanty@gatech.edu}{ayush.mohanty@gatech.edu}.}
}
\affil{Georgia Institute of Technology, Atlanta, GA 30332, USA}

\author{
Xiaolei Fang\footnote{Associate Professor, Edward P. Fitts Department of Industrial and Systems Engineering.}
}
\affil{North Carolina State University, Raleigh, NC 27607, USA}

\author{
Stephen K. Robinson\footnote{Professor, Department of Mechanical and Aerospace Engineering.}
}
\affil{University of California, Davis, Davis, CA 95616, USA}

\author{
Nagi Gebraeel\footnote{Coca-Cola Foundation Chair and Professor, H. Milton Stewart School of Industrial and Systems Engineering.}
}
\affil{Georgia Institute of Technology, Atlanta, GA 30332, USA}
\begin{document}
\maketitle
%% Abstract
\begin{abstract}

\subfile{Abstract.tex}
\end{abstract}

% \linenumbers

%% Nomenclature
\clearpage
\section*{Nomenclature}
{\renewcommand\arraystretch{1.0}
\noindent\begin{longtable*}{@{}l @{\quad=\quad} l@{}}
ASGL & Adaptive Sparse Group Lasso \\
CA & Covariate-Adjusted \\
CES & Complex Engineering System \\
% CNSA & China National Space Administration \\
C-MAPSS & Commercial Modular Aero-Propulsion System Simulation \\
CDLL & Complete-Data Log-Likelihood \\
DSH & Deep-Space Habitat \\
ECLSS & Environmental Control and Life Support System \\
% EKF & Extended Kalman Filter \\
EM & Expectation-Maximization \\
% ESA & European Space Agency \\
\hl{FMEA} & \hl{Failure Mode and Effect Analysis} \\
FPCA & Functional Principal Component Analysis \\
FVE & Fraction of Variance Explained \\
\hl{HPC} & \hl{High Pressure Compressor} \\ 
IDLL & Incomplete-Data Log-Likelihood \\
% ILRS & International Lunar Research Station \\
% ISS & International Space Station \\
% JAXA & Japan Aerospace Exploration Agency \\
KNN & K-Nearest Neighbors \\
MFPCA & Multivariate Functional Principal Component Analysis \\
MGR & Mixture of Gaussian Regressions \\
NASA & National Aeronautics and Space Administration \\
% PCA & Principal Component Analysis \\
PDF & Probability Density Function \\
RUL & Remaining Useful Life \\
SNR & Signal-to-Noise Ratio \\
TTF & Time-to-Failure \\
\end{longtable*}}

\section{Introduction}
\subfile{Introduction.tex}

\section{Literature Review}\label{lit}
\subfile{LitReview.tex}

\section{Overview Of Our Prognostics Framework}\label{PF}
\subfile{Overview.tex}

\section{Offline Sensor Selection}\label{SS}
\subfile{OfflineSensorSelection.tex}
\section{Online Diagnosis and Prognostics}\label{DDP}
\subfile{OnlineDiagnosisPrognostics.tex}

\section{Case Study 1: Simulated Dataset}\label{Sim}
\subfile{CaseStudy1.tex}

\section{Case Study 2: NASA C-MAPSS Turbofan Engine Dataset}\label{Real}
\subfile{CaseStudy2.tex}

\section{Conclusion}\label{Conclusion}
\subfile{Conclusion.tex}

\section*{Acknowledgments}
This effort is supported by NASA under grant number 80NSSC19K1052 as part of the NASA Space Technology Research Institute (STRI) Habitats Optimized for Missions of Exploration (HOME) ‘SmartHab’ Project. Any opinions, findings, conclusions, or recommendations expressed in this material are those of the authors and do not necessarily reflect the views of the National Aeronautics and Space Administration.

\appendix
\subfile{Appendix.tex}



%\newpage
\bibliography{ref}

% %% For numbered reference style

%   \textit{\LaTeX: a document preparation system},
%   Addison Wesley, Massachusetts,

% \end{thebibliography}
\end{document}

%% file: Abstract.tex
Deep-space habitats (DSHs) are safety-critical systems that must operate autonomously for long periods, often beyond the reach of ground-based maintenance or expert intervention. Monitoring system health and anticipating failures are therefore essential. Prognostics based on remaining useful life (RUL) prediction support this goal by estimating how long a subsystem can operate before failure. Critical DSH subsystems, including environmental control and life support, power generation, and thermal control, are monitored by many sensors and can degrade through multiple failure modes. These failure modes are often unknown, and informative sensors may vary across modes, making accurate RUL prediction challenging when historical failure data are unlabeled.
We propose an unsupervised prognostics framework for RUL prediction that jointly identifies latent failure modes and selects informative sensors using unlabeled run-to-failure data. The framework consists of two phases: an offline phase, where system failure times are modeled using a mixture of Gaussian regressions and an Expectation–Maximization algorithm to cluster degradation trajectories and select mode-specific sensors, and an online phase for real-time diagnosis and RUL prediction using low-dimensional features and a weighted functional regression model. The approach is validated on simulated DSH telemetry data and the NASA C-MAPSS benchmark, \hl{demonstrating its ability to identify unknown failure modes, select mode-specific informative sensors, and accurately predict RUL.}

%% file: Introduction.tex
\hl{In 2017, NASA initiated the later-named Artemis program \cite{lane2025probabilistic, creech2022artemis} to resume human lunar exploration and develop infrastructure for future deep-space operations. A key element of NASA’s original Artemis program was the lunar Gateway, a deep space habitat (DSH) that would orbit the Moon to support lunar exploration and serve as a testbed for technological development required for future human missions to Mars  \cite{kirshner2022integrating}. The Gateway was designed to operate autonomously for extended crewless periods, requiring sophisticated health monitoring and control software beyond that traditionally seen in human spaceflight \cite{fuller2022gateway, hollaway2020eyes}. Beyond NASA, space agencies worldwide are independently pursuing DSH infrastructure. The European Space Agency (ESA) and the Japan Aerospace Exploration Agency (JAXA) are jointly developing the Lunar I-Hab module for the Gateway, a pressurized habitat designed to support crews during lunar orbital missions and to operate autonomously during extended crewless periods \cite{fuller2022gateway}. In parallel, the China National Space Administration (CNSA) is planning a phased lunar surface habitat through its International Lunar Research Station (ILRS) program, with a basic model near the lunar south pole targeted for the mid-2030s \cite{wu2023international}. These concurrent multinational programs collectively underscore that autonomous health management of DSHs is a critical, globally recognized challenge for the coming decades of human space exploration \cite{gratius2024digital, carbone2023fault}.

A DSH is a complex engineering system (CES) composed of tightly integrated subsystems that must operate autonomously to sustain human life and achieve mission objectives over extended exploration campaigns. Critical subsystems such as the environmental control and life support system (ECLSS), electrical power systems, and thermal management system must function reliably in harsh and isolated environments \cite{des2021deep, zaccarine2024monitoring}. Component failures can degrade the performance of these subsystems, compromising both the completion of mission objectives and the lives of people onboard. Accurate prediction of a system’s remaining useful life (RUL), the time between the current operation time and the future system failure time, is an important prognostic capability because it supports long-term operability of DSHs by providing advanced warning of failure \cite{yin2024time}. These warnings could then inform appropriate maintenance actions \cite{rozas2025data}.

% Critical subsystems such as the environmental control and life support system (ECLSS), energy and power systems, thermal control, robotic agents, and structural components must function reliably in harsh and isolated environments, where failure of any component can lead to mission-critical consequences \cite{gordon2024improving, zaccarine2024monitoring}. %For example, in the ECLSS alone, failures can originate from membrane degradation in the water recovery system, contaminant buildup in the trace contaminant control assembly, or pump bearing wear, each producing distinct sensor signatures across overlapping sensor sets \cite{ibrahim2025generative, rautela2023real}. 
% Ensuring the long-term operability of DSHs, therefore, requires %not only fault detection but also 
% accurate prediction of remaining useful life (RUL), defined as the difference between the current operation time and the future system failure time \cite{yin2024time, sharma2024risk}.

RUL prediction in DSHs is challenging because DSH systems may experience multiple failure modes. For example, a failure mode and effects analysis (FMEA) of an oxygen generation system identified the electrolysis unit, separation device, and pump, amongst other components, as potential sources of failure \cite{guo2020nonlinear}. Even within the electrolysis unit, two separate failure modes were identified: the decline of current efficiency and membrane electrode performance degradation \cite{guo2020nonlinear}. For systems with multiple failure modes, RUL can depend on the active failure mode. Thus, accurate RUL prediction may require an initial diagnosis of this failure mode \cite{Xiong2023}. This task is made more challenging when failure-mode labels are unavailable, which can occur when the physics of the degradation process is not well-understood or when accurate information about the cause of failure is not available \cite{cofre2021big}. In an example of the latter, a report by Oeftering et al. \cite{oeftering2011component} analyzing the International Space Station’s (ISS’s) Problem Reporting and Corrective Action database noted that the root causes of a significant number of Electric Power System faults were labeled unknown. Thus, a historical repository of observed DSH failures may not contain complete failure mode labels.

DSHs are expected to be equipped with an extensive fleet of sensors for monitoring the state-of-health of its systems. For context, NASA’s Habitat Demonstration Unit \cite{howe2013nasa} was equipped with over 100 sensors by the end of 2011. These sensors included accelerometers and temperature, pressure, and humidity sensors to monitor subsystems such as thermal control, environmental control and life support, and vegetable production. Similarly, the International Space Station (ISS) is monitored by approximately 350,000 sensors, and a few of the next-generation DSHs are expected to operate at comparable sensor densities \cite{iverson2012general, rollock2022defining}. The multivariate time series data generated from such arrangements can aid in RUL prediction because the signal trajectories offer a rich source of information from which features related to system health can be extracted. However, there may exist redundancy between sensors, where multiple sensors capture the same physical phenomena. Furthermore, some sensors may offer no information for predicting RUL. Thus, selecting a subset of informative sensors for health monitoring can result in improved RUL prediction accuracy \cite{Fang2017}. This task can become more challenging with multiple failure modes because the characteristics of signal trajectories (rate of change, direction of change, and amplitude at system failure) may depend on the failure mode \cite{Xiong2023}. When failure modes are unknown, there exists an interdependency between failure mode identification and sensor selection (Figure \ref{fig:motivation}). Recent prognostics (i.e., RUL prediction) approaches have acknowledged this interdependency but still assume failure mode labels are known a priori. This paper presents a prognostics framework that uses multivariate sensor data with \textbf{no failure mode labels} to \textbf{jointly} identify latent failure modes and select a subset of informative sensors for each failure mode.

\textbf{Main Contributions.} The framework presented in this paper comprises two phases. In the first phase, we train a Mixture of Gaussian Regression with Adaptive Sparse Group Lasso (MGR-ASGL) model to jointly perform failure mode identification and sensor selection. This phase is performed offline either onboard the DSH or on Earth. The second phase is designed to be performed autonomously. This phase leverages the inferred failure mode labels and the selected sensors to learn two models: a diagnostic model and a prognostic model. The diagnostic model utilizes the sensors selected across all failure modes to diagnose the active failure mode of a fielded system. Then, the prognostic model utilizes the sensors selected for the diagnosed failure mode to predict its RUL. This phase can be performed in an autonomous, adaptive fashion, periodically refitting the diagnostic and prognostic model in real-time as new sensor data becomes available. The key technical contributions of our framework can be summarized as follows:}
\begin{itemize}
    \item \hl{We develop a joint failure mode identification and sensor selection methodology that simultaneously identifies latent failure modes from historical degradation-to-failure data and selects sensors informative for each identified failure mode, without requiring known failure mode labels.}
    \item \hl{We develop a feature extraction methodology that utilizes Covariate-Adjusted Functional Principal Component Analysis (CA-FPCA) to extract prognostic features that account for the heterogeneity in signal trajectories.}
    \item \hl{We develop an autonomous online diagnostic and prognostic methodology for fielded assets. This methodology fuses the trajectories from the selected sensors into informative features that are used to (1) diagnose the active failure mode and (2) predict RUL using a mode-specific regression model.}
    \item \hl{We evaluate the performance of our proposed methodology using two case studies: (1) a controlled simulation designed to reflect the challenges of performing prognostics for a CES such as multiple unknown failure modes, a large number of sensors, and signals corrupted by noise, and (2) the NASA C-MAPSS turbofan engine degradation dataset, exhibiting telemetry data similar to a degrading DSH system.}
\end{itemize}

The remainder of the paper is organized as follows. Section \ref{lit} reviews related work on RUL prediction and multi-failure mode diagnostics. Section \ref{PF} introduces our proposed prognostics framework. Section \ref{SS} describes the offline sensor selection method. Section \ref{DDP} details the online diagnosis and RUL prediction approach. Section \ref{Sim} and Section \ref{Real} present two case studies, a simulated dataset and the NASA turbofan engine dataset, respectively. We conclude the paper in Section \ref{Conclusion} with a discussion of future work.

\begin{figure}[H]
    \centering
    \includegraphics[scale=0.325]{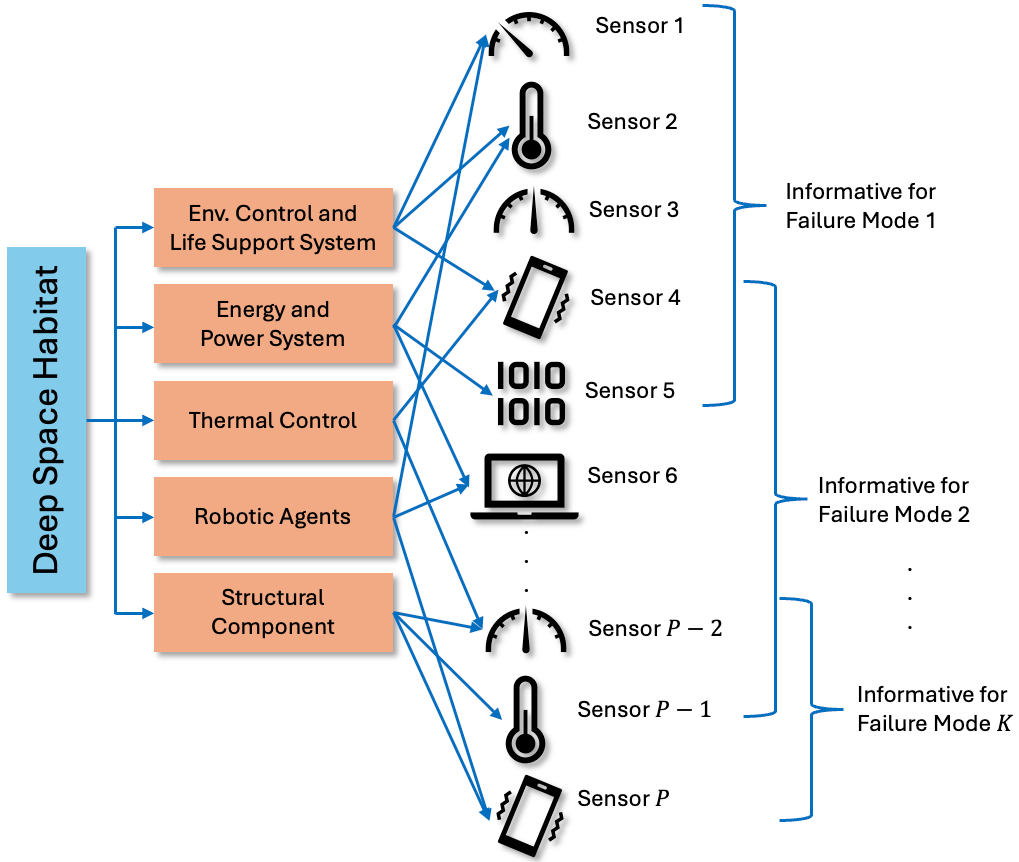}
    \caption{Example of a deep-space habitat with multiple subsystems. The health of each subsystem is monitored by a subset of onboard sensors. The subset of informative sensors varies for different failure modes.}
    \label{fig:motivation}
\end{figure}

%% file: LitReview.tex
Ensuring the operability of DSHs over extended autonomous missions requires an accurate prediction of RUL. The RUL prediction problem has been approached in the broader engineering literature through either physics-based or data-driven methods. Physics-based approaches \cite{mirfarah2025estimation, jin2019physics} utilize first principles to mathematically model the underlying physics of system degradation. For example, \cite{mirfarah2025estimation} used a two-stage nonlinear differential equation to model the rate of air pressure changes inside a space habitat. They use an Extended Kalman Filter (EKF) to estimate the parameters of this model and predict RUL by using the same EKF to forecast when the habitat pressure becomes untenable. Meanwhile, \cite{jin2019physics} integrated a physics degradation model with a Gaussian process to predict RUL in rollerball bearings. In general, physics-based approaches typically develop state-space models of system degradation and use a Bayesian filter such as the Kalman Filter or the Particle Filter for real-time estimation and prediction of model parameters. Several examples of this are highlighted by \cite{kordestani2023overview}. In the context of DSHs, physics-based approaches are particularly challenging to develop given the novel failure mechanisms induced by microgravity, radiation, and thermal extremes that have no terrestrial analogs \cite{rollock2022defining, zaccarine2024monitoring}. The advantages of physics-based approaches include high prediction accuracy and limited need for data. However, their requirement for thorough knowledge of the physics behind system failure renders them difficult to develop, given growing system complexity.

Conversely, data-driven approaches attempt to map sensor data to the system RUL using statistical models, machine learning, or hybrid approaches. Since sensors are monitoring the system over a prolonged period, they generate continuous, high-volume data called "streaming data." The approaches developed for single-sensor streaming data include both statistical methods that model univariate degradation signals using exponential models \cite{gebraeel2006sensory, zhang2020remaining}, random-coefficient regression \cite{wang2014real}, Brownian motion \cite{si2012remaining}, Gamma process \cite{rodriguez2018degradation}, or hidden Markov model \cite{wang2007prognosis} and machine learning methods that use deep learning \cite{pei2022bayesian, kumari2024efficient} or relevance vector machine \cite{zhang2021data}. However, single-sensor approaches are not scalable to the scenario where multiple sensors monitor systems. As an example, \cite{Satoh2024} mounted 51 sensors on a reusable rocket engine. Meanwhile, the International Space Station is monitored by approximately 350,000 sensors \cite{rollock2022defining}, and next-generation DSHs such as the lunar Gateway are expected to operate at comparable sensor densities during extended crewless periods \cite{badger2024integrated}, generating high-volume multivariate streaming data that single-sensor approaches cannot handle. For this type of data, authors have developed sensor fusion algorithms that operate at either the decision level or the data level \cite{Song2019}. Decision-level fusion consists of systematically combining RUL estimates from different models, whereas data-level fusion synthesizes the multivariate sensor data into a univariate health indicator, which can be modeled using standard univariate techniques. Existing data fusion approaches include Principal Component Analysis \cite{Wen2019, Gu2021, Fang2017}, maximum likelihood methods \cite{Kim2019}, extreme learning machine \cite{lu2019aircraft}, deep learning \cite{Wang2022, che2019combining}, state-space models \cite{Li2021, Wu2022, Daroogheh2017}, and logistic regression \cite{yu2017aircraft}.

While effective, many data fusion methods rely on aggregating all available sensor signals. In practice, not all sensors are informative for a given failure mechanism, and fusing irrelevant signals can introduce noise and reduce prediction accuracy. To address this, some studies incorporate sensor selection. For instance, adaptive lasso has been used to eliminate uninformative sensors \cite{Song2019, Kim2019}, and group regularization techniques have been applied for joint selection and modeling \cite{Fang2017}. However, these approaches typically assume a single dominant failure mode. This single-mode assumption is particularly limiting for DSHs, where subsystems such as the ECLSS, power generation, and thermal control can each degrade through distinct and overlapping failure mechanisms \cite{ibrahim2025generative, zaccarine2024monitoring}.

Given their complex nature, DSHs will eventually experience multiple failure modes. Competing risk models have been widely used to represent such scenarios. Traditional approaches often rely on statistical models, such as the Cox proportional hazards model with Weibull baselines \cite{Zhang2014, Cox1972}; a comprehensive review is provided in \cite{Monterrubio2022}. More recently, deep learning has also been used to model competing risks \cite{Zhu2016}. For example, both \cite{Gupta2019} and \cite{Marthin2023} employ deep recurrent neural networks for recurrent event survival analysis with competing risks. Competing risk models often produce survival or hazard rates that can be difficult to interpret in engineering contexts. For a DSH operating autonomously far from Earth, the inability to interpret hazard rates in engineering terms is especially problematic, as maintenance decisions must be made onboard without expert intervention \cite{rollock2025characterizing, ulusoy2025investment}. As a result, many researchers have shifted toward methods that directly model sensor data to predict RUL, though few focus on multiple failure modes. Often, researchers develop general RUL prediction methodologies and then apply them to datasets with multiple failure modes. For example, \cite{Aremu2020} proposed a correlation and relative entropy feature engineering framework for complex systems and \cite{Kim2021} proposed a deep convolutional neural network (CNN). In both cases, the researchers validated their methodologies on datasets with multiple failure modes. However, they do not consider multiple failure modes in an explicit manner. Failure to do so could result in inaccurate predictions, which could prove critical in a deep-space environment.

Recently, the development of RUL prediction methodologies that account for multiple failure modes has grown. \cite{Jiao2020} used a support vector machine for failure mode diagnosis with features extracted from a deep belief network, followed by RUL prediction using a particle filter. \cite{Xiong2023} trained distinct LSTM models for each failure mode. Once a physics-informed CNN classifier diagnoses the active failure mode, the appropriate LSTM model is utilized. Furthermore, \cite{Chehade2018} fit random coefficient models to failure mode-specific health indices. However, these articles do not incorporate a sensor selection algorithm and they rely on labeling of the failure modes determined either through visual inspection or domain knowledge. However, DSHs must operate autonomously and thus cannot rely on failure mode labeling from an operator on Earth. The time to complete the labeling process coupled with the distance from Earth to the DSH could result in significant and unacceptable delays \cite{rollock2025characterizing}.

Consequently, researchers have begun addressing these gaps. \cite{Wu2023} adopted a semi-supervised graph-based approach for feature extraction from partially labeled failures and used elastic net functional regression for sensor selection and RUL prediction. More recently, \cite{Fu2025} proposed a time-series clustering algorithm to identify latent failure modes, followed by LSTM-based RUL prediction. Furthermore, \cite{su2025deep} combined a mixture of (log)- location-scale regression with deep learning to predict RUL in a scenario with unknown failure modes. However, in both \cite{Fu2025} and \cite{su2025deep}, sensor selection was limited to excluding sensors with obviously non-informative signals, such as constant or low-variance signals. Fitting a model with data from too many sensors may result in overfitting, which can reduce the prediction accuracy of the model. This problem would be exacerbated by the sheer volume of sensors mounted on a DSH. Therefore, our methodology differs from prior work by addressing a setting in which failure mode labels are entirely unknown. We assume that each failure mode is associated with a (not necessarily disjoint) subset of sensors that are informative for RUL prediction. The goal is to develop an unsupervised prognostic framework that systematically fuses multivariate sensor signals for RUL estimation. We assume access to a historical repository of system run-to-failure trajectories, where degradation was monitored from the onset of an incipient fault to final failure using multiple sensors. Furthermore, we assume that each system fails independently and experiences only one failure mode during its lifetime. Finally, while the failure labels are unobserved, we assume that the total number of potential failure modes is known. This problem is particularly acute for DSHs, where the combination of high sensor counts \cite{zhang2025managing}, unknown failure modes \cite{zaccarine2024monitoring}, and communication delays that preclude expert labeling \cite{rollock2025characterizing} collectively demands an unsupervised, sensor-selective prognostic framework of the kind proposed in this paper. Table~\ref{tab:literature_comparison} summarizes the key differences
between existing prognostic methodologies and our proposed framework. In the following sections, we detail our prognostics framework.

\setcounter{table}{0}
\begin{table}
\centering
\caption{Comparison of Prognostic Methodologies in the Literature}\label{tab:literature_comparison}
\begin{adjustbox}{max width=\columnwidth}
\begin{tabular}{lccccc}
\hline
\textbf{Prognostic} &
\textbf{Sensor} &
\textbf{Sensor} &
\textbf{Multiple} &
\textbf{Unlabeled} &
\textbf{Model} \\
\textbf{Methodology} & \textbf{Fusion} & \textbf{Selection} &
\textbf{Failure Modes} & \textbf{Failure Modes} &
\textbf{Interpretability}\\ 
\hline

\cite{mirfarah2025estimation, jin2019physics, kordestani2023overview}
& \xmark & \xmark & \xmark & N/A & \cmark \\

\cite{gebraeel2006sensory, zhang2020remaining, wang2014real,
si2012remaining, rodriguez2018degradation, wang2007prognosis}
& \xmark & \xmark & \xmark & \xmark & \cmark \\

\cite{Wen2019, Gu2021, Fang2017, Kim2019, Wang2022}
& \cmark & \xmark & \xmark & \xmark & Partial \\

\cite{Kim2019, Fang2017, Song2019}
& \cmark & \cmark & \xmark & \xmark & \cmark \\

\cite{Chehade2018, Jiao2020, Xiong2023}
& \cmark & Partial & \cmark & \xmark & \xmark \\

\cite{Wu2023}
& \cmark & \cmark & \cmark & Partial & Partial \\

\cite{Fu2025, su2025deep}
& \cmark & Partial & \cmark & \cmark & Partial \\

\cite{Zhu2016, Gupta2019, Marthin2023}
& \cmark & \xmark & \cmark & \xmark & \xmark \\

\cite{Aremu2020, Kim2021}
& \cmark & \xmark & \cmark & \xmark & \xmark \\

\textbf{Ours}
& \cmark & \cmark & \cmark & \cmark & \cmark \\

\hline
\end{tabular}
\end{adjustbox}
\end{table}

%% file: Overview.tex
\begin{figure}[h]
    \centering
    \includegraphics[width=\columnwidth]{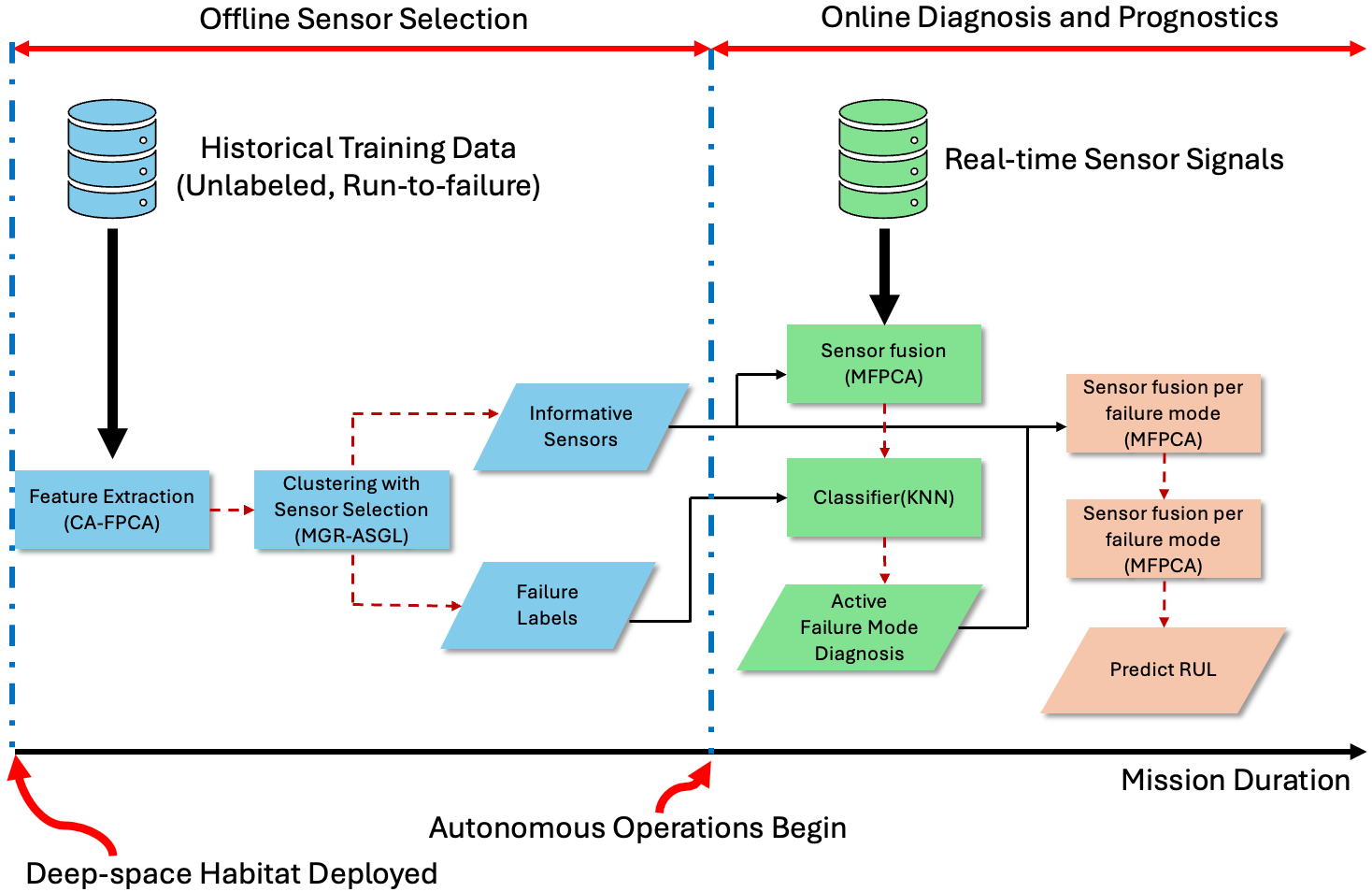}
    \caption{Overview of the proposed prognostics framework for deep-space habitats}
    \label{fig:method}
\end{figure}

This section describes our prognostics framework, developed for autonomous health monitoring in deep space habitats. A flowchart of the framework is shown in Figure~\ref{fig:method}. The framework consists of two main components: ``\textit{Offline Sensor Selection}'' and ``\textit{Online Diagnosis and Prognostics}''.

\textbf{I. Offline Sensor Selection:} This stage occurs shortly after the habitat is deployed, during an initial period where degradation signals are collected with limited ground or crew assistance. The data collected during this stage forms a historical dataset, which is used to initialize the prognostic models before autonomous operation begins. We assume that the dataset contains multiple failure modes, although the \textbf{\textit{failure modes are not labeled}}. The goal of this step is to \textbf{(1)} label the degradation signals into one of the failure modes, and \textbf{(2)} identify a subset of informative sensors for each failure mode. This step consists of two parts, discussed below:

\begin{enumerate}
   \item \textbf{Feature Extraction:} The onboard sensors generate high-dimensional, time-varying signals. To extract informative features for prognostics, we use a covariate-adjusted functional principal component analysis (CA-FPCA) method. CA-FPCA models the variation in each sensor signal while accounting for external covariates. In our setting, these covariates represent the underlying failure modes, which are unknown a priori. To estimate them, we first perform a K-means clustering on the sensor signals and use the resulting cluster labels as initial covariates. 

    \item \textbf{Optimization:} After feature extraction, we obtain a reduced set of features for each sensor. These features are called CA-FPC scores. We use these scores as predictors in a mixture of Gaussian regression (MGR) model. The MGR model estimates the natural logarithm of the time-to-failure, $\ln{TTF}$, as a function of the CA-FPC scores. The model assumes that this relationship ($\ln{TTF}$ vs CA-FPC scores) depends on the underlying failure mode. We fit the MGR model using the EM algorithm, which results in: (1) \textbf{\textit{optimal failure mode labels}} for each sample, and (2) an \textbf{\textit{optimal subset of informative sensors}} for each failure mode.
\end{enumerate}

\textbf{II. Online Diagnosis and Prognostics:} Once the failure mode labeling and sensor selection are complete, the habitat enters into autonomous mode. This stage uses real-time sensor data to \textbf{(1)} identify the active failure mode and \textbf{(2)} predict the RUL: 
\begin{enumerate}
    \item \textbf{Diagnosis:} Prior to the diagnosis step, the active failure mode is unknown. Therefore, all selected sensors are utilized in this step. First, we apply multivariate functional principal component analysis (MFPCA) to signals from all selected sensors to extract a compact set of features, called MFPC scores, that represent the joint behavior of all sensors. Then we classify the active failure mode by finding its K-nearest neighbors and assigning the most common failure mode among them.
    \item \textbf{Prognostics:} Once the active failure mode is diagnosed, we recalculate the MFPC scores using only the informative sensors for that mode. A functional regression model uses these scores to predict the RUL.
\end{enumerate}

%% file: OfflineSensorSelection.tex
We consider a scenario where the DSH is monitored by $P$ different sensors. We assume there exists a historical dataset comprised of $P$-dimensional degradation-based sensor signals from monitoring $N$ similar systems. We denote $s_{i,p}(t)$ as the signal data from sensor $p\in\{1,...,P\}$ for system $i\in\{1,...,N\}$ observed at time $t\in[0,T]$, where $T=min(\{TTF_i\}_{i=1}^N)$. By truncating all signals to the minimum failure time, we ensure that all signals in the training set have the same time domain and that they are all included when modeling. For this framework, we assume that the total number of failure modes, $K$ is known, but that the actual failure labels in the training dataset are unknown. This poses a challenge for CA-FPCA, which requires observed covariates for computation. To determine these covariates, we perform a sensor-wise clustering of the training samples. For each sensor, we perform FPCA for feature extraction and then use K-means clustering with $K$ clusters to label the signals from that sensor. Once the sensor-wise clustering is complete, the next step involves feature extraction by applying CA-FPCA and then regressing these features against the time-to-failure to identify the most informative subset of sensors for each failure mode. The next two subsections discuss details about the feature extraction (CA-FPCA), and the optimization (MGR-ASGL) model.

\subsection{Feature Extraction: CA-FPCA}\label{CAFPCA}
CA-FPCA extends conventional FPCA to handle complex data structures where the dynamics of time-varying signals are affected by one or more additional factors, i.e. covariates. In \cite{Jiang2010}, the covariates are assumed to be continuous variables, and thus, mean and covariance estimation requires smoothing over both time and the covariate space. In this paper, \textbf{we use the sensor-wise cluster labels as covariates}. Let $\mathcal{I}_p^k$ denote the set of indices for sensor $p$ signals assigned to cluster $k, k\in\{1,2,...,K\}$. The CA-FPCA problem was reduced to performing FPCA on each cluster. For a given system $i$, signals from cluster $k$ of sensor $p$ are modeled as:
\begin{equation}\label{eq6_Ng}
s_{i, p}(t)=v_{i,p}^k(t)+\epsilon_{i,p}(t), i\in\mathcal{I}_k^p
\end{equation}\label{eq6}
\noindent where $v_{i, p}^k(t)$ is a smooth random function, and $\epsilon_{i, p}(t)$ are assumed to be independently and identically distributed (i.i.d.) errors with mean zero and variance $\varsigma_{p}^2$. $v_{i,p}^k(t)$ and $\epsilon_{i,p}(t)$ are assumed to be independent of each other. The mean and covariance functions of $v_{i,p}^k(t)$ are given by $\mu_p^k(t)$ and $\mathbb{C}_p^k(t, t^{\prime})$, respectively. Using Mercer's theorem \cite{Karhunen1947}, the covariance can be decomposed as follows, $\mathbb{C}_p^k(t, t^{\prime})=\sum_{m=1}^{\infty}\lambda_{m,p}^k\phi_{m,p}^k(t) \phi_{m,p}^k(t^{\prime})$, where $\{\phi_{m,p}^k(t)\}_{m=1}^\infty$ are the orthogonal eigenfunctions of $\mathbb{C}_p^k(t,t^{\prime})$ and $\lambda_{1,p}^k\geq...\geq\lambda_{m,p}^k\geq...$ are the corresponding eigenvalues. By projecting the mean-subtracted signal data onto the eigenfunctions, we can represent the signals by Eq (\ref{eq6}).
\vspace*{-1em}
\begin{equation}\label{eq6}
    s_{i, p}(t)=\mu_p^k(t)+\sum_{m=1}^{\infty} \xi_{i, m, p}^k \phi_{m, p}^k(t)+\epsilon_{i, p}(t)
\end{equation}
where $\xi_{i,m,p}^k$ is the $m$th CA-FPC-score with mean 0 and variance $\lambda_{m,p}^k$. The CA-FPC scores are computed using the PACE algorithm in \cite{Yao2005}.
Since the eigenvalues decrease as $m \to \infty$, we can obtain a low-dimensional representation of $s_{i,p}^k(t)$ using the first $q_p$ CA-FPC-scores.

To ensure all features are scaled equally, we standardize the CA-FPC-scores by subtracting the sample mean and dividing by the sample standard deviation. Let $x_{i,m}^p$ denote the standardized $m$th CA-FPC-score from sensor $p$ of system $i$. We extract $\boldsymbol{x}_{i,p}=(x_{i,1}^p,...,x_{i,q_p}^p)^T$ for sensor $p$ of system $i$,. We then utilize the $\boldsymbol {x}_{i,p}$'s as predictors for fitting the MGR-ASGL model.

\subsection{Optimization of the MGR-ASGL model}\label{MGRASGL}
Let $Y_i=\frac{\ln{TTF}_i-(1/N)\sum_{i=1}^N \ln{TTF}_i}{(1/(N-1))\sum_{i=1}^N\big(\ln{TTF}_i-(1/N)\sum_{i=1}^N \ln{TTF}_i\big)^2}$, for $i=1,2…,N$ denote the random variable corresponding to the standardized natural logarithm of the time-to-failure (i.e., $\ln{TTF}$) of system $i$. Furthermore, let $\boldsymbol{Z}_i$ denote a categorical random vector of size $K$ used to encode the unknown failure mode of system $i$. If failure mode $k$ is responsible for the failure of system $i$, which occurs with probability $\pi_k$, then $\boldsymbol{Z}_i[k]=1$ and the remaining elements are zero. In our model, we posit that the relationship between $Y_i$ and the observed features $x_i$ depends on which failure mode caused the failure of system $i$. Thus, we assume that the conditional distribution of $Y_i$ given $\boldsymbol{Z}_i[k]=1$ is Gaussian with PDF:
\begin{equation}
     f_{Y_i|\boldsymbol{Z}_i}(y_i|\boldsymbol{Z}_i[k]=1)=\frac{1}{\sqrt{2\pi}\sigma_k}\exp\Big(-\frac{1}{2\sigma_k^2}(y_i-\beta_{0,k}-\sum_{p=1}^P \boldsymbol{x}_{i,p}^T\boldsymbol{\beta}_{p,k})^2\Big)
\end{equation}
To find the marginal distribution of $Y_i$ we use the law of total probability, summing over the joint distribution of $Y_i$ and $\boldsymbol{Z}_i$ as follows:
\begin{equation}
      f_{Y_i}(y_i)=\sum_{k=1}^K \pi_k\frac{1}{\sqrt{2\pi}\sigma_k}\exp\Big(-\frac{1}{2\sigma_k^2}(y_i-\beta_{0,k}-\sum_{p=1}^P \boldsymbol{x}_{i,p}^T\boldsymbol{\beta}_{p,k})^2\Big) 
\end{equation}
Therefore, the marginal distribution of $Y_i$ is a mixture of Gaussian regressions. The parameters of this model are $\Theta=\{\pi_k,\beta_{0,k},\boldsymbol{\beta}_{1,k},\dots,\boldsymbol{\beta}_{P,k},\sigma_k\}_{k=1}^K$. To estimate these parameters, we seek to minimize the negative incomplete-data log-likelihood (IDLL). To ensure both scale-invariance of parameter estimation and convexity of the optimization problem in the M-step of our EM algorithm, we let $\varphi_{0,k}=\sigma_k^{-1}\beta_{0,k}$, $\boldsymbol{\varphi}_{p,k}=\sigma_k^{-1}\boldsymbol{\beta}_{p,k}$, and $\rho_k=\sigma_k^{-1}$ (see: [41]). Thus, the parameters are now $\Theta=\{\boldsymbol{\pi},\boldsymbol{\rho},\boldsymbol{\varphi}\}$, where $\boldsymbol{\pi}=\{\pi_{k}\}_{k=1}^K$, $\boldsymbol{\rho}=\{\rho_k\}_{k=1}^K$, and $\boldsymbol{\varphi}=\{\varphi_{0,k},\boldsymbol{\varphi}_{1,k},\dots,\boldsymbol{\varphi}_{P,k}\}_{k=1}^K$. Assuming that $Y_1,\dots,Y_N$ are independent, the negative IDLL is computed as follows:
\begin{equation}
    -\ell(\Theta|\boldsymbol{Y})=-\sum_{i=1}^N\ln\Big[\sum_{k=1}^K \pi_k  \rho_k\frac{1}{\sqrt{2\pi}}\exp\Big(-\frac{1}{2}(y_i\rho_k-\varphi_{0,k}-\sum_{p=1}^P \boldsymbol{x}_{i,p}^T\boldsymbol{\varphi}_{p,k})^2\Big)\Big]
\end{equation}
Estimating $\Theta$ requires minimizing Eq 5. However, the logarithm inside the summation makes Eq 5 difficult to optimize globally. Instead, we search for a local minimum using the EM algorithm, an iterative algorithm suitable for fitting probability models with latent variables. To utilize the EM algorithm, we first construct the negative complete-data log likelihood (CDLL).
\begin{equation}
\begin{split}
    -\ell_C&(\Theta|\boldsymbol{Y},\mathbb{Z})=-\ln\bigg(\prod_{i=1}^N\prod_{k=1}^K f_{Y_i}(y_i)^{\boldsymbol{Z}_{i}[k]}\bigg)\\&=-\sum_{i=1}^N\sum_{k=1}^K \boldsymbol{Z}_{i}[k]\ln\Big[\pi_k\rho_k\frac{1}{\sqrt{2\pi}}\exp\Big(-\frac{1}{2}(y_i\rho_k-\varphi_{0,k}-\sum_{p=1}^P \boldsymbol{x}_{i,p}^T\boldsymbol{\varphi}_{p,k})^2\Big)\Big]
\end{split}
\end{equation}
In ~\ref{app1}, we show that the expectation of the negative CDLL with respect to the conditional distribution of $\mathbb{Z}$ given $\boldsymbol{Y}$ forms an upper bound for the negative IDLL. Therefore, we can obtain a local minimum by minimizing this upper bound. The proposed EM algorithm is performed as follows:
\begin{enumerate}
    \item \textbf{Initialization}: Set the initial parameters to $\Theta^{(0)}$ s.t., 
    \begin{equation}
        \Theta^{(0)} = \{\pi_k^{(0)},\varphi_{0,k}^{(0)},\boldsymbol{\varphi}_{1,k}^{(0)},\dots,\boldsymbol{\varphi}_{P,k}^{(0)},\rho_k^{(0)}\}_{k=1}^K
    \end{equation}
    \item \textbf{E-step}: Given $\Theta^{(m)}$, the expectation of $-\ell_C(\Theta|\boldsymbol{Y},\mathbb{Z})$ is given as: 
    \begin{equation}
    \begin{split}
        Q&(\Theta,\Theta^{(m)})=\mathbb{E}[-\ell_C(\Theta|\boldsymbol{Y},\mathbb{Z})|\boldsymbol{Y},\Theta^{(m)}]\\&=-\sum_{i=1}^N\sum_{k=1}^K \gamma_{i,k}\ln\Big[\pi_k\rho_k\frac{1}{\sqrt{2\pi}}\exp\Big(-\frac{1}{2}(y_i\rho_k-\varphi_{0,k}-\sum_{p=1}^P \boldsymbol{x}_{i,p}^T\boldsymbol{\varphi}_{p,k})^2\Big)\Big]
    \end{split}
    \end{equation}
\begin{equation}
\begin{split}
\text{where,} \hspace{0.5cm} \gamma_{i,k}
&:=
\mathbb{E}\!\left(\boldsymbol{Z}_{i}[k]\mid Y_i=y_i,\Theta^{(m)}\right)
\\
&=
\Pr\!\left(\boldsymbol{Z}_{i}[k]=1 \mid Y_i=y_i,\Theta^{(m)}\right)
\\
&=
\begingroup
\footnotesize
\frac{
\begin{aligned}
 \pi_k^{(m)}\rho_k^{(m)}\frac{1}{\sqrt{2\pi}}
 \exp\!\Big(-\frac{1}{2}
 \big(
   y_i\rho_k^{(m)}-\varphi_{0,k}^{(m)}
   -\sum_{p=1}^P
   \boldsymbol{x}_{i,p}^T\boldsymbol{\varphi}_{p,k}^{(m)}
 \big)^2\Big)
\end{aligned}
}{
\begin{aligned}
 \sum_{l=1}^K
 \pi_l^{(m)}\rho_l^{(m)}\frac{1}{\sqrt{2\pi}}
 \exp\!\Big(-\frac{1}{2}
 \big(
   y_i\rho_l^{(m)}-\varphi_{0,l}^{(m)}
   -\sum_{p=1}^P
   \boldsymbol{x}_{i,p}^T\boldsymbol{\varphi}_{p,l}^{(m)}
 \big)^2\Big)
\end{aligned}
}
\endgroup
\end{split}
\end{equation}

$\gamma_{i,k}$ is the responsibility that failure mode $k$ takes for the failure of system $i$. These responsibilities enable a soft clustering, where we claim system $i$ failed due to failure mode $k$ if $\gamma_{i,k}>\gamma_{i,l}$ for $l\neq k$.
\item \textbf{M-Step}: After computing the responsibilities $\gamma_{i,k}$ , we can update our estimate of $\Theta$ by maximizing $Q(\Theta,\Theta^{(m)})$ with respect to $\Theta$. To perform sensor selection, we augment $Q(\Theta,\Theta^{(m)})$ with the adaptive sparse group lasso penalty and form the following optimization problem:
\begin{equation}
     \min_{\Theta}\Big\{Q(\Theta,\Theta^{(m)})+\lambda\sum_{k=1}^K \pi_k\Big(\alpha\sum_{p=1}^P\|\boldsymbol{\varphi}_{p,k}\|_1+(1-\alpha)\sum_{p=1}^P\sqrt{q_p}\|\boldsymbol{\varphi}_{p,k}\|_2\Big)\Big\}
\end{equation}
We refer to the problem in Eq 10 as Mixture of Gaussian Regression–Adaptive Sparse Group Lasso (MGR-ASGL). Because $\boldsymbol{\pi}$ is included in the penalty, we follow \cite{Stadler2010} and use a two-step optimization:

\begin{itemize}
    \item \textbf{\textit{Step 1: Improvement w.r.t. $\boldsymbol{\pi}$}}

We start by using the responsibilities from the E-step to identify a feasible solution \\$\bar{\boldsymbol{\pi}}^{(m+1)}=\big(\frac{1}{N}\sum_{i=1}^N \gamma_{i,1},\dots,\frac{1}{N}\sum_{i=1}^N \gamma_{i,K}\big)^T$. Then, we update the mixing coefficients as follows:
\begin{equation}
    \boldsymbol{\pi}^{(m+1)}=\boldsymbol{\pi}^{(m)}+u_m(\bar{\boldsymbol{\pi}}^{(m+1)}-\boldsymbol{\pi}^{(m)})
\end{equation}
where $u_m\in(0,1]$ is chosen to be the largest value in the sequence $\{1,\frac{1}{2},\frac{1}{4},\dots\}$ such that the objective function does not increase.
\item \textbf{\textit{Step 2: Improvement w.r.t. $\boldsymbol{\rho}$ and $\boldsymbol{\varphi}$}}

Given the updated mixing coefficients, we can optimize Eq 10 with respect to the other parameters. It can be shown that this problem can be decomposed into solving $K$ individual problems, one for each failure mode. Therefore, for $k=1,\dots,K$, we solve
\begin{equation}
\begin{split}
    \min \Big\{&-\sum_{i=1}^N \gamma_{i,k}\big[\ln\pi_k^{(m+1)}+\ln\rho_k-\frac{1}{2}(y_i\rho_k-\varphi_{0,k}-\sum_{p=1}^P \boldsymbol{x}_{i,p}^T\boldsymbol{\varphi}_{p,k})^2\big]\ \\&+\lambda\pi_k^{(m+1)}\Big(\alpha\sum_{p=1}^P\|\boldsymbol{\varphi}_{p,k}\|_1+(1-\alpha)\sum_{p=1}^P\sqrt{q_p}\|\boldsymbol{\varphi}_{p,k}\|_2\Big)\Big\}
\end{split}
\end{equation}
Note that the tuning parameters $\lambda$ and $\alpha$ do not change for different $k$. These parameters can be selected using cross-validation. When $\lambda$ increases, ASGL drives groups of regression coefficients to zero. Since these groups correspond to features extracted from individual sensors, this, in effect, causes the influence of some sensors to vanish. Therefore, a sensor is considered significant for failure mode $k$ if $\|\varphi_{p,k}\|_2\neq 0$. The parameter $\alpha$ allows for some of the coefficients in a group to drop to zero to improve prediction accuracy. Finally, incorporating the adaptive parameters $q_p$ and $\gamma_{i,k}$ into the penalty ensures that sensor selection is not affected by class imbalance and the dimensionality of the features. Let $\boldsymbol{\rho}^{(m+1)}$ and $\boldsymbol{\varphi}^{(m+1)}$ denote the solution to Eq 12. Thus, after completing Steps 1 and 2, we have our updated solution $\Theta^{(m+1)}=\{\boldsymbol{\pi}^{(m+1)},\boldsymbol{\rho}^{(m+1)},\boldsymbol{\varphi}^{(m+1)}\}$.
\end{itemize}
Set $\Theta^{(m+1)}\leftarrow \Theta$ and return to \textbf{(1)}. Perform successive iterations of \textbf{(1)}–\textbf{(3)} until convergence.
\end{enumerate}

%% file: OnlineDiagnosisPrognostics.tex
% This section details the steps of the online part of our multi-failure mode prognostics framework. 
In this section, a system within a deep-space habitat is considered to be degrading in real-time. Our goal is to predict the RUL of the system. We begin by showing how to fuse the signals from multiple sensors into a parsimonious set of features. Then, we demonstrate how we utilize the fused sensor information to i) diagnose the active failure mode of the degrading system in real-time, and ii) predict its RUL.
% \begin{algorithm}[h]
% \caption{Online Diagnosis and Prognostics} \label{alg:algo2}
% \SetKwInOut{Input}{Input}
% \SetKwInOut{Output}{Output}

% \Input{Smoothed signals from training \& test units, Failure labels of training data from Algorithm \ref{alg:algo1}, Set of informative sensors $\mathcal{P}_k$ from Algorithm \ref{alg:algo1}}
% \Output{Active failure mode in the test unit $k^{*}$, Remaining Useful Life (RUL) of the test unit $RUL_{test}$}

% \vspace{0.1cm}
% \tcc{Diagnostic Model Training}
% $\zeta_{(i), h} \gets$ MFPCA on training units (Eq. \ref{eq14})\;
% Train a KNN classifier using $\zeta_{(i), h}$\;

% \vspace{0.2cm}
% \tcc{Diagnostic Model Testing}
% $\zeta_{test, h} \gets$ MFPCA on test unit\;
% $k^* \gets$ Mode of labels in KNN of $\zeta_{test, h}$\;

% \vspace{0.3cm}
% \tcc{Prognostic Model Training}
% $\zeta^{k^*}_{(i), h} \gets$ MFPCA: selected sensors (Eq. \ref{eq18})\;
% $c_0^{k^*}, \dots, c_h^{k^*} \gets$ Obtain $\argmin$ in Eq. \ref{eq20}\;

% \vspace{0.2cm}
% \tcc{Prognostic Model Testing}
% $\zeta^{k^*}_{test, h} \gets$ MFPCA on test unit (Eq. \ref{eq18})\;
% $RUL_{test} \gets$ Eq \ref{eq22}\;
% \end{algorithm}

\subsection{Sensor Fusion: MFPCA}
Without loss of generality, let us redefine the sensor index as $\{1,2, \ldots, \mathcal{P}\}$, where $\mathcal{P}$ is number of sensors we seek to fuse. 
Now assume that we have monitored the degrading system for a duration given by the time interval $[0,t^*]$. Since we are interested in RUL, we only use signals from training systems with failure times greater than $t^*$. We define this set as the ``$t^*-$training'' set, whose indices we denote by the set $\mathcal{I}_{t^*}$, which has cardinality $N_{t^*}$. Furthermore, we smooth the degradation signals using the 'rloess' method as described in \cite{Cleveland2012}. This method uses weighted least squares with a 2nd-order polynomial for smoothing. The weights come from a bisquare weight function requiring a specified bandwidth parameter. This method also identifies outliers and assigns them weights of zero to improve signal mean estimation. To perform MFPCA, we model the $i$th smoothed multivariate signal evaluated at time $t\in[0,t^*]$ as having mean $\boldsymbol{\mu}(t)\in\mathcal{R}^\mathcal{P}$ and $\mathcal{P}\times\mathcal{P}$ block covariance function $\mathcal{C}(t,t^{\prime})$, where $\mathcal{C}_{j,j^{\prime}}(t,t^{\prime})=cov(v_j(t),v_{j^{\prime}}(t^{\prime}))$. Using Mercer's theorem \cite{Karhunen1947}, we can represent the covariance function as $\mathcal{C}(t,t^{\prime})=\sum_{h=1}^\infty\eta_h\boldsymbol{\varpi}_h(t)\boldsymbol{\varpi}_h(t^{\prime})^T$, where $(\eta_h,\boldsymbol{\varpi}_h(t))$ are the $h$th eigenvalue-eigenfunction pair and $\eta_1\geq\eta_2\geq\dots$. Using the first $H$ multivariate eigenfunctions, we can represent the multivariate signal by Eq \ref{eq14} where the MFPC-scores $\zeta$'s are computed according to Eq \ref{eq15}.
\begin{equation}\label{eq14}
    \boldsymbol{s}_{i}(t)=\boldsymbol{\mu}(t)+\sum_{h=1}^{H} \zeta_{i, h} \boldsymbol{\varpi}_h(t)+\boldsymbol{\epsilon}_{i}(t), \vspace*{-1.2em}
\end{equation}
\begin{equation}\label{eq15}
    \zeta_{i,h}=\int_{0}^{t^*}(\boldsymbol{s}_{i}(t)-\boldsymbol{\mu}(t))^T\boldsymbol{\varpi}_{h}(t)dt
\end{equation}
In practice, we perform MFPCA by collecting the $N_{t^*}$ observations, each consisting of the $\mathcal{P}$ sensor signals concatenated into a single vector. Then we perform traditional PCA on the sample of observations. Furthermore, we standardize the extracted MFPC-scores so they have zero mean and unit variance.
\subsection{Failure Mode Diagnosis: K-nearest Neighbors}
In order to diagnose the failure mode of a degrading system at time $t^*$, 
we apply MFPCA on the $ t^*$ training dataset. Next, K-nearest neighbors (KNN) is applied to the standardized MFPC-scores to identify the active failure mode. Here, the distance metric of the KNN classifier is defined using the Euclidean distance between the MFPC-scores. Given this diagnosis, we now develop a model for predicting system RUL.

\subsection{RUL Prediction: Weighted Functional Regression}\label{FunctionalRegression}
To predict RUL, we use a weighted time-varying functional regression model that maps multivariate signal features to the $\ln{TTF}$. First, we discuss the basics of a time-varying functional regression and then motivate the case for a weighted version of the model
\subsubsection{Time-Varying Functional Regression}
When predicting RUL, we are often faced with updating the relationship between time-to-failure and the current values of the extracted signal features (MFPC-scores). The predictor trajectories up to the current time are represented by time-varying MFPC-scores, which are continuously updated as time progresses and are considered to be time-varying predictor variables for the functional regression model. This is referred to as \textit{time-varying functional regression}. It was first proposed by \cite{muller2005time} and has been used for prognostics in \cite{Fang2015, Fang2017}. In our problem setting, let $k^*$ denote the diagnosed failure mode and $\mathcal{I}_{t^*}^{k^*}\subset\mathcal{I}_{t^*}$ denote the set of indices for training systems that failed due to failure mode $k^*$, but survived up to time $t^*$. Furthermore, let $\mathcal{P}_{k^*}\subset\{1,2,...,\mathcal{P}\}$ denote the set of sensors selected (identified as most informative) for failure mode $k^*$. To predict RUL, we fit a functional regression model as shown in Eq (\ref{eq16}), where $y_{i}=\ln{TTF_{i}}$, $\varphi_{0}^{k^*}$ is the bias term, $\boldsymbol{\varphi}^{k^*}(t)\in\mathcal{R}^{|\mathcal{P}_{k^*}|}$ is the multivariate coefficient function, and $\boldsymbol{s}_{i}^{k^*}(t)$ is the multivariate signal consisting only of sensors significant to failure mode $k^*$. 
\begin{equation}\label{eq16}
    \min _{\varphi_{0}^{k^*},\boldsymbol{\varphi}^{k^*}(t)} \sum_{i\in\mathcal{I}_{t^*}^{k^*}}\left(y_{i}-\varphi_0-\int_{0}^{t^*}\boldsymbol{\varphi}^{k^*}(t)^T\boldsymbol{s}_{i}^{k^*}(t)dt\right)^2
\end{equation}
We expand the coefficient function using the eigenfunctions obtained from applying MFPCA on the training set $\mathcal{I}_{t^*}^{k^*}$. The regression coefficient function can now  be expanded as shown in Eq (\ref{eq17}), where $H_{k^*}$ denotes the number of MFPC-scores that were retained for failure mode $k^*$. 
\begin{equation}\label{eq17}
    \boldsymbol{\varphi}(t)=\sum_{h=1}^{H_{k^*}} c_{h}^{k^*} \boldsymbol{\varpi}_{k}^{k^*}(t)
\end{equation}
Likewise, we expand the multivariate signals using the eigenfunctions as shown in Eq (\ref{eq18}). 
\begin{equation}\label{eq18}
    \boldsymbol{s}_{i}^{k^*}(t)=\boldsymbol{\mu}^{k^*}(t)+\sum_{h=1}^{H_{k^*}} \zeta_{i, h}^{k^*} \boldsymbol{\varpi}_h^{k^*}(t)
\end{equation}
Since the eigenfunctions are orthogonal, substituting Eqs (\ref{eq17}) and (\ref{eq18}) into Eq (\ref{eq16}) yields the following optimization problem, 
\begin{equation}\label{eq19}
    \min _{c_0^{k^*}, c_1^{k^*}, \ldots, c_{H_{k^*}}^{k^*}} \sum_{i\in\mathcal{I}_{t^*}^{k^*}}\left(y_{i}-c_0^{k^*}-\sum_{h=1}^{H_{k^*}} \zeta_{i, h}^{k^*} c_h^{k^*}\right)^2
\end{equation}
where, \begin{equation}
c_{0}^{k^*}=\varphi_{0}^{k^*}+\int_{0}^{t^*} \sum_{h=1}^{H_{k^*}} c_{h}^{k^*} \boldsymbol{\varpi}_{h}^{k^*}(t)^T \boldsymbol{\mu}^{k^*}(t)
\end{equation}

\textbf{Weighted Version:}
The offline sensor selection algorithm is unlikely to cluster failures perfectly. Therefore, when fitting the regression model, it is possible that sample systems that failed due to other failure modes are included. These systems may present themselves as outliers, compromising RUL prediction accuracy. To address this issue, we use weighted least squares to fit the time-varying functional regression model. To derive the weights, we first measure the Euclidean distance between the MFPC score of each training system to the centroid of the cluster of the diagnosed failure mode. Then we calculate the reciprocal of that distance to define the corresponding weights for each system in the training set. If a training system's failure mode is misclassified, its corresponding weight $w_{i}$ would be large, and thus, training system $i$ would be assigned a smaller weight in the regression problem. Next, we regress the standardized $\ln{TTF}$ on the MFPC-scores using a penalized weighted linear regression shown in Eq (\ref{eq20}). For variable selection, we include the lasso penalty in the optimization problem. To ensure a regression coefficient represents the significance of their respective predictor to the prediction of the failure time, Eq (20) should be read such that $\zeta_{i,h}^{k^*}$ is standardized. 
\begin{equation}\label{eq20}
    \min _{c_0^{k^*}, c_1^{k^*}, \ldots, c_{H_{k^*}}^{k^*}} \sum_{i\in\mathcal{I}_{t^*}^{k^*}}\left(\sqrt{w_{i}}y_{i}-\sqrt{w_{i}}c_0^{k^*}-\sum_{h=1}^{H_{k^*}}\sqrt{w_{i}}\zeta_{i, h}^{k^*} c_{h}^{k^*}\right)^2 +\lambda\sum_{h=1}^{H_{k^*}}
    \lVert c_{h}^{k^*} \rVert
\end{equation}
\subsubsection{RUL prediction}
As we have shown, MFPCA enables the transformation of the functional regression problem into a multiple linear regression problem, which allows us to solve the problem by estimating only $H_{k^*}+1$ parameters. After fitting the regression coefficients, we extract the MFPC-scores for the test signal $\zeta_{test, 1}^{k^*}, \ldots, \zeta_{test, H_{k^*}}^{k^*}$. These MFPC scores are standardized using the sample mean and sample standard deviation of the MFPC scores from the training set. These scores are then used to predict the RUL at time $t^*$ using Eq (\ref{eq22}).
\begin{equation}\label{eq22}
\begin{split}
{RUL}_{test}=\exp \bigg(\Big[c_0^{k^*}&+\sum_{k=1}^{H_{k^*}} \zeta_{test, h}^{k^*} c_k^{k^*}\Big]\times\Big[(1/(N-1))\sum_{i=1}^N\big(\ln{TTF}_i-(1/N)\\&\times\sum_{i=1}^N \ln{TTF}_i\big)^2\Big]+(1/N)\sum_{i=1}^N\ln{TTF_i} \bigg) -t^*
\end{split}
\end{equation}

%% file: CaseStudy1.tex
\hl{In this section, we present a simulation study to evaluate the efficacy of our proposed framework. This study adapts the simulation in \cite{Fang2017} to generate datasets with telemetry data consistent with the scenario described in Figure \ref{fig:motivation}. We emphasize that the equations used in this simulation are mathematical abstractions for generating degradation signals rather than physics-based equations that describe the dynamics of specific failure processes in a real system. In this simulation, we generate three datasets with varying levels of signal noise. We use these datasets to evaluate the robustness of our proposed framework in recovering the latent failure modes, selecting sensors important for prognostics, and predicting remaining useful life under different data quality conditions.}

\subsection{Simulation Model}
\begin{figure}
    \centering
  \subfloat[\label{1a}]{%
       \includegraphics[width=0.25\textwidth]{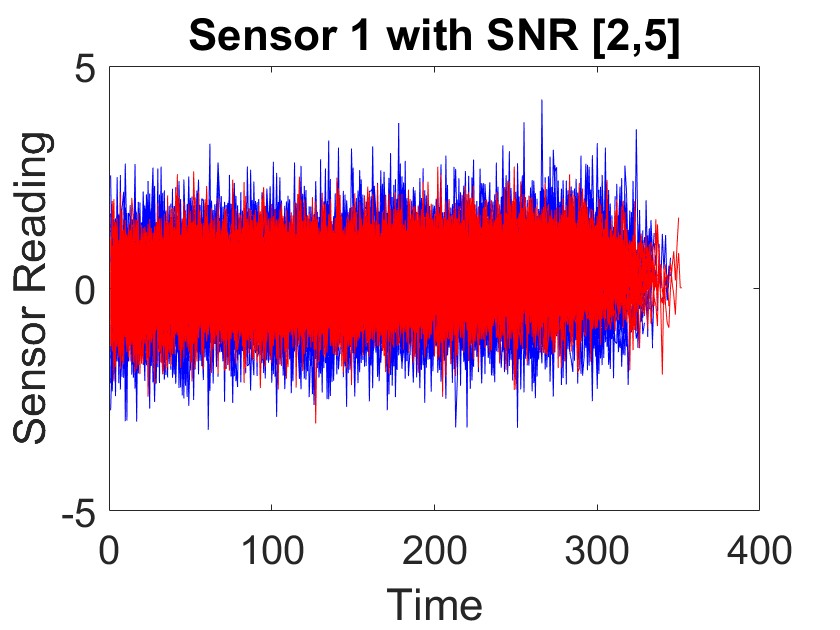}}\hfil
  \subfloat[\label{1b}]{%
        \includegraphics[width=0.25\textwidth]{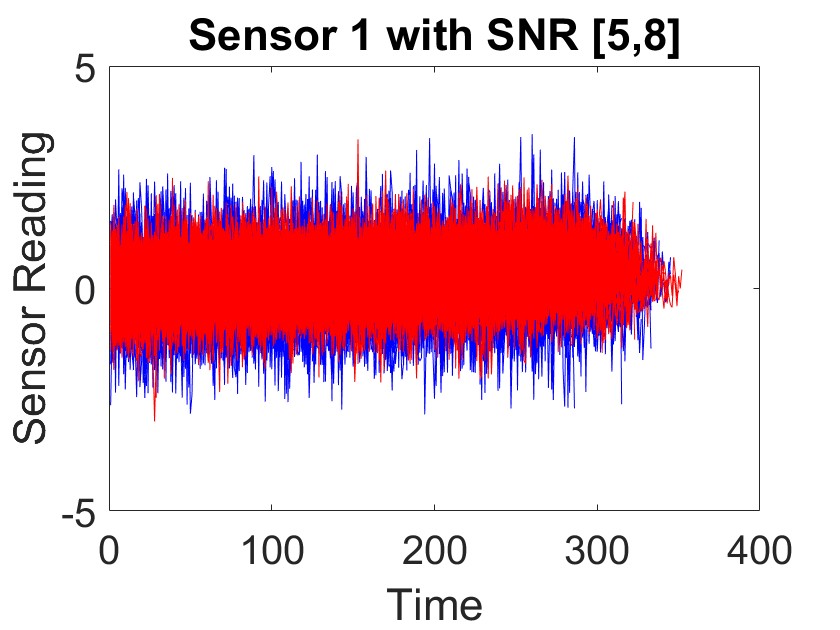}}
               \hfil
      \subfloat[\label{1c}]{%
       \includegraphics[width=0.25\textwidth]{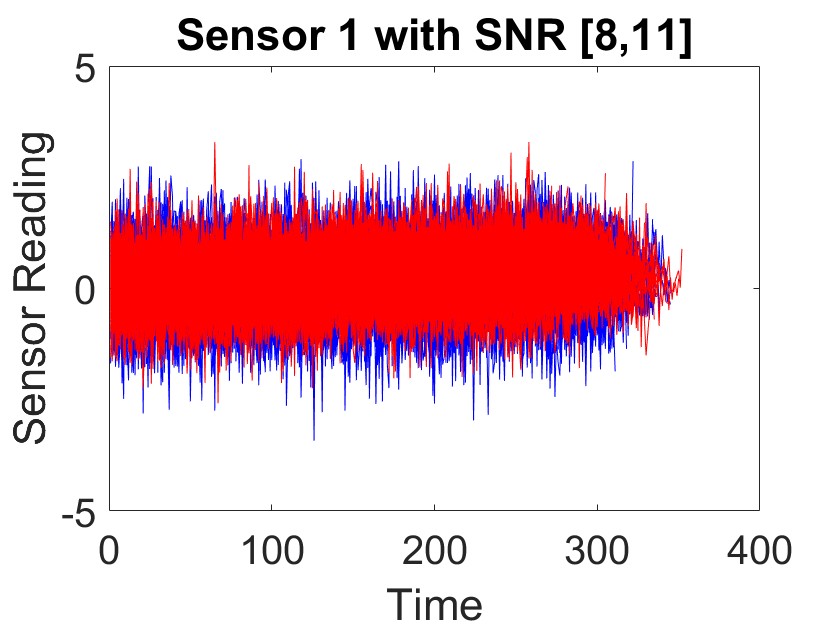}}\\
  \subfloat[\label{1d}]{%
        \includegraphics[width=0.25\textwidth]{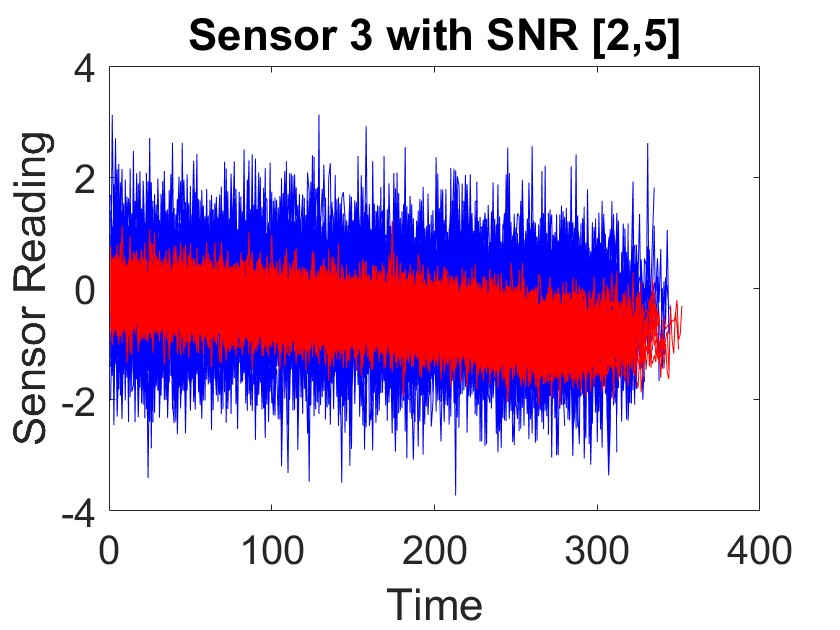}}\hfil
    \subfloat[\label{1e}]{%
    \includegraphics[width=0.25\textwidth]{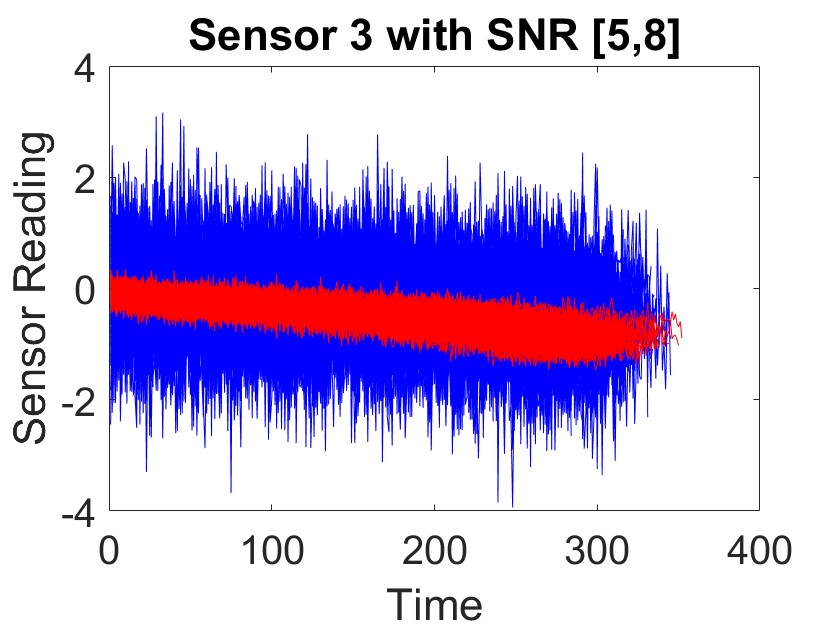}}\hfil
  \subfloat[\label{1f}]{%
        \includegraphics[width=0.25\textwidth]{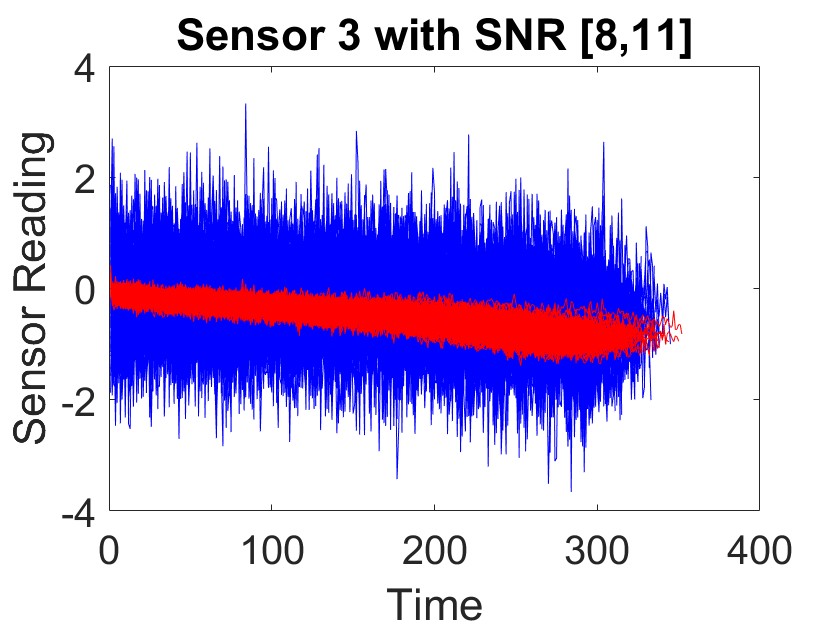}}\\
    \subfloat[\label{1g}]{%
       \includegraphics[width=0.25\textwidth]{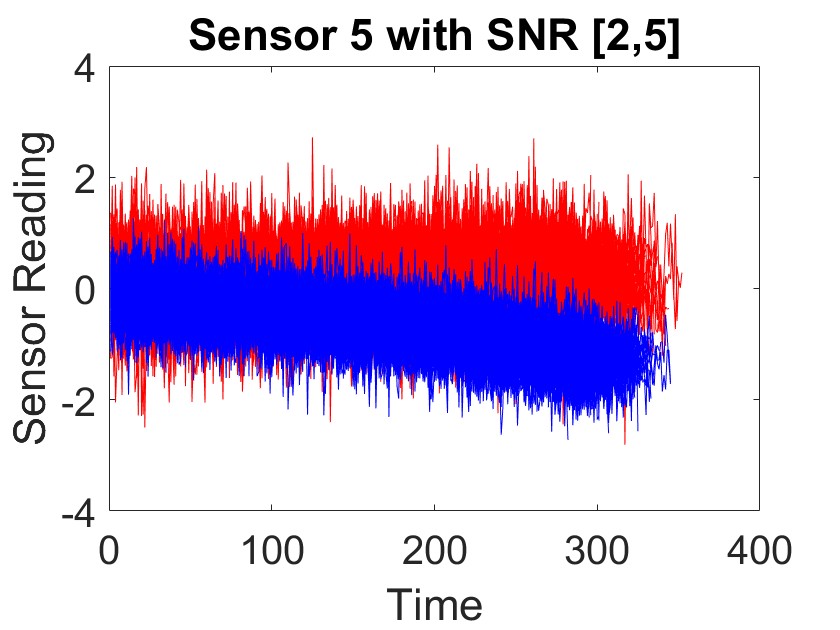}}\hfil
  \subfloat[\label{1h}]{%
        \includegraphics[width=0.25\textwidth]{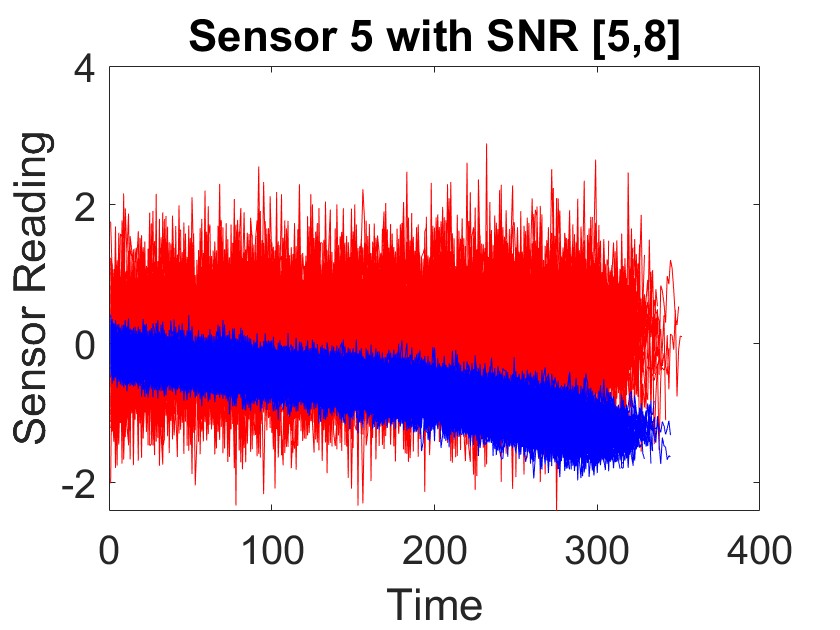}}\hfil 
     \subfloat[\label{1i}]{%    
        \includegraphics[width=0.25\textwidth]{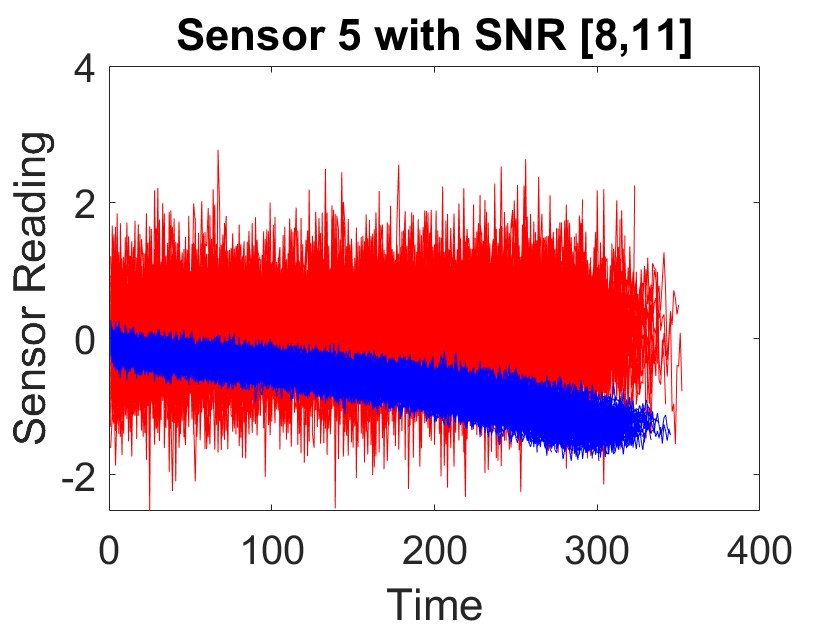}}\\
  \subfloat[\label{1j}]{%
        \includegraphics[width=0.25\textwidth]{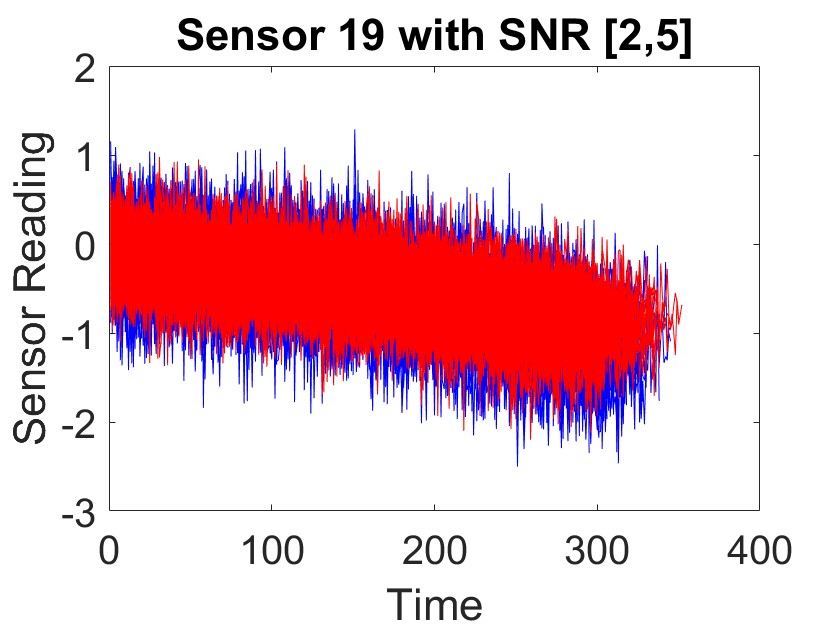}}
               \hfil
\subfloat[\label{1k}]{%
       \includegraphics[width=0.25\textwidth]{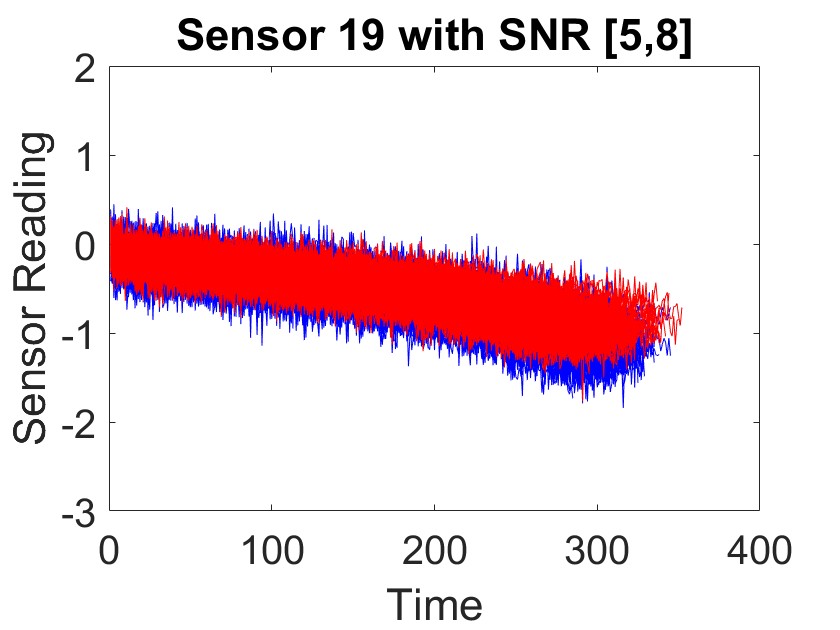}}\hfil
  \subfloat[\label{1l}]{%
        \includegraphics[width=0.25\textwidth]{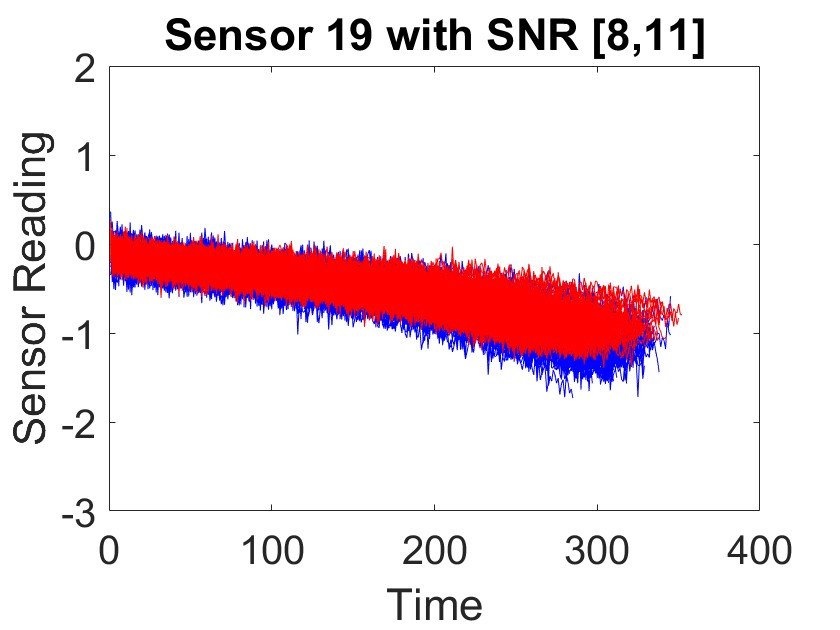}}\\
  \caption{\hl{Case study 1 data: Signals in sensors 1, 3, 5, 19 (rows 1, 2, 3, 4 respectively) for FM 1 (blue) and FM 2 (red) under different SNRs: [2, 5], [5, 8], and [8, 11] (columns 1, 2, 3 respectively)}}
  \label{fig:F1sensors} 
\end{figure}

\hl{For this simulation, we assume a system can fail due to one of two failure modes: FM 1 or FM 2. For each failure mode, we generated 200 realizations of degradation-to-failure data. The latent degradation state at time $t$ of the $i$th system that fails due to FM $k$ is governed by Eq \ref{eqSimulate1},}
\begin{equation}\label{eqSimulate1}
s_i^k(t)=-\frac{\theta_i^k}{\ln t}
\end{equation}
\hl{where $\theta_i^k \sim N(\mu_k,0.1^2)$ represents the growth rate of the latent degradation state. We set $\mu_1$ equal to 1.0 and $\mu_2$ equal to 0.8. Furthermore, we define the time to failure of the $i$th system that fails due to FM $k$ as $TTF_i^k=\inf(t:s_i^k(t)\ge D_k)$, the first time the latent degradation state passes the failure threshold $D_k$. In our simulation, we set $D_1=2$ and $D_2=1.5$. By using a unique threshold for each failure mode, we simulate a scenario in which the latent degradation state at failure varies between failure modes. This allows us to simulate a scenario such as one component failing at a lower temperature than another component.

After we simulated the latent degradation state for each system, we generated the $P$ sensor signals observed during degradation. As described in Figure \ref{fig:motivation}, each failure mode may have a subset of sensors that are highly responsive to the underlying system degradation and are therefore strongly informative for predicting RUL. The subsets associated with different failure modes may also share common sensors. In addition, other sensors may exhibit little responsiveness to the underlying system degradation and are therefore only weakly informative for predicting RUL. In our simulation, we considered a system monitored by $P=20$ sensors. For FM 1, the set of strongly informative sensors was $\mathcal{I}_1=\{5,12,16,19\}$, whereas the set of strongly informative sensors for FM 2 was $\mathcal{I}_2=\{{3,7,9,19}\}$ For FM $k$, the complementary set $\mathcal{I}_k^c$ contained the remaining 16 weakly informative sensors. To generate the $p$th sensor signal for the $i$th system exhibiting FM $k$, we utilize Eq (\ref{eqSimulate2}),}
\begin{equation}\label{eqSimulate2}
    s_{i, p}^k(t)=-\frac{\theta_{i, p}^k}{\ln t}+\epsilon_{i, p}^k(t)
\end{equation}
\hl{where $\theta_{i,p}^k$ is the growth (decay) rate of the signal from sensor $p$ when FM $k$ is present. For a strongly informative sensor, we sampled this parameter from the conditional distribution $\theta_{i, p}^k \mid \theta_i^k \sim 
N\left(\mu_k\left(1-\sqrt{1-\rho_p^k}\right),(0.1)^2\right)$, where we drew $\rho_p^k=CORR(\theta_i^k,\theta_{i,p}^k)$ randomly from the interval $[0.80,0.99]$. For $p\in\mathcal{I}_k^c$, we draw $\rho_p^k$ randomly from the interval $[0.10,0.60]$ By specifying the conditional mean in Eq. (\ref{eqSimulate2}) as a function of the latent degradation rate in Eq. (\ref{eqSimulate1}), we generated strongly informative sensor signals whose growth (decay) rates were highly correlated with the underlying system degradation rate. In contrast, the weakly informative sensor signals were less strongly associated with the underlying degradation process and approached a white noise process as $\rho_p^k$ approached 0. To generate signals that could either increase or decrease with degradation, we randomly multiplied $\{\theta_{i,p}^k,i=1,...,N\}$ by either 1 or -1.

To ensure that the underlying degradation processes remained constant across signal-to-noise (SNR) conditions, we used the same generated growth (decay) rates for all three datasets. However, we varied the error process $\epsilon_{i,p}^k(t)$ for each dataset. $\epsilon_{i,p}^k(t) \mid N(0,(\sigma_k^p)^2))$ is a white noise process parameterized by $\sigma_k^p=\mu_k/SNR_p^k$, where $SNR_p^k$ represents the SNR for the sensor-failure mode pair $(p,k)$. For $p\in\mathcal{I}_k$, we drew $SNR_p^k$ randomly from $[2,5]$, $[5,8]$, and $[8,11]$ for datasets 1, 2, and 3 respectively. Conversely, for $p\in\mathcal{I}_k^c$, we drew $SNR_p^k$ randomly from $[1,3]$. Figure \ref{fig:F1sensors} displays samples of our sensor signals in a 4$\times$3 grid. Each column displays signals from four sensors $(1, 3, 5, 19)$ at a particular SNR. Sensor 1 is weakly informative for both failure modes. Therefore, its signals are very noisy and exhibit a slight increasing trend over time. Sensor 3 is strongly informative for FM 2, whereas Sensor 5 is strongly informative for FM 1. For these sensors, the signals associated with the active failure mode exhibit a decaying trajectory in response to degradation, while the remaining signals exhibit noisy trajectories similar to those in Sensor 1. Finally, Sensor 19 is strongly informative for both failure modes, and its signals decay in response to degradation under both FM 1 and FM 2.}

\subsection{Evaluation of Proposed Framework}

\hl{In this subsection, we describe the application of our proposed framework to the simulated datasets and evaluate its performance over each SNR scenario.}

%In this subsection, we evaluate the clustering and sensor selection capability of the MGR-ASGL model. After generating the telemetry data, we randomly partitioned the data into a training set (320 systems) and a test set (80 systems). Using the training set, we utilized CA-FPCA as described in Section \ref{CAFPCA} to extract features for each sensor separately. To determine the number of CA-FPC scores to retain, we used the 95\% fraction of variance explained (FVE) criterion. For each within-sensor cluster, we selected the first $q_{p}^k$ CA-FPC scores whose normalized eigenvalues sum to at least 0.95. Since the number of CA-FPC scores to retain can vary for each cluster, we retain  $q_p=\max_k{q_p^k}$ CA-FPC scores. %Thus, we refer to the CA-FPC scores of system $i$ as $\boldsymbol{\xi}_i=(\boldsymbol{\xi}_{i,1}^T,...,\boldsymbol{\xi}_{i,20}^T)$, where $\boldsymbol{\xi}_{i,p}^T\in\mathbb{R}^{q_p}$.
\subsubsection{Data Splitting and Phase 1 Feature Extraction}
\hl{Each dataset is comprised of 200 samples from both FM 1 and FM 2. We randomly partitioned these samples into a training set with 160 samples per failure mode and a test set with 40 samples per failure mode. To ensure a valid comparison, we held the sample indices in the training set and test set constant for each dataset.

In Phase 1, we extracted features from each sensor using CA-FPCA as described in Section \ref{CAFPCA}. For sensor $p$, we determined the number of CA-FPC scores to retain using the 95\% fraction of variance explained (FVE) criterion. First, we used K-means clustering with two clusters to partition the FPC scores. Then, we used the cluster index as a covariate for CA-FPCA. Within cluster k, we selected the first $q_p^k$ CA-FPC scores whose normalized eigenvalues sum to at least 0.95. Since the number of CA-FPC scores to retain tend to differ for each cluster, we retained $q_p=max_k{1_p^k}$ CA-FPC scores for sensor $p$. We then repeated this for all sensors.}
\subsubsection{MGR-ASGL Model Fitting}
\begin{figure}
    \centering
  \subfloat[\label{a}]{%
       \includegraphics[width=0.3\textwidth]{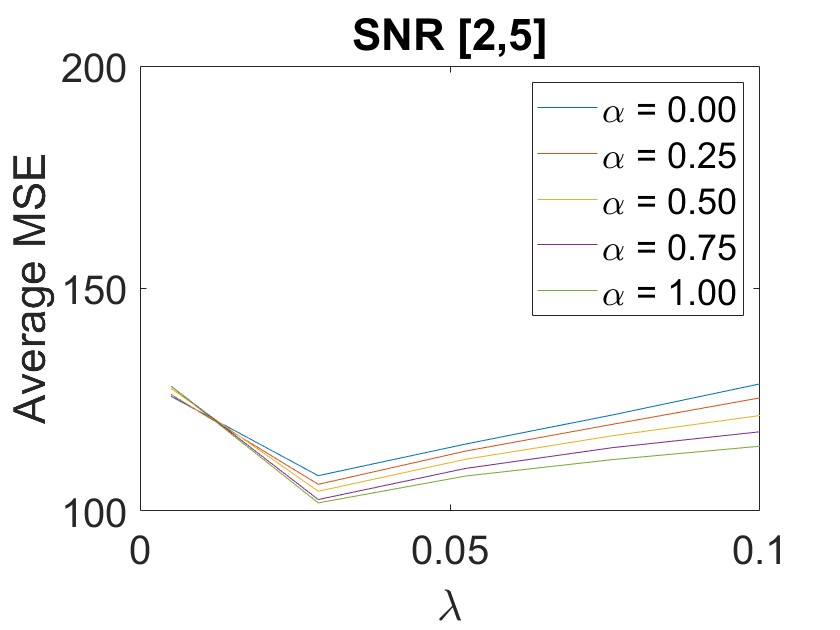}}\hfil
  \subfloat[\label{b}]{%
        \includegraphics[width=0.3\textwidth]{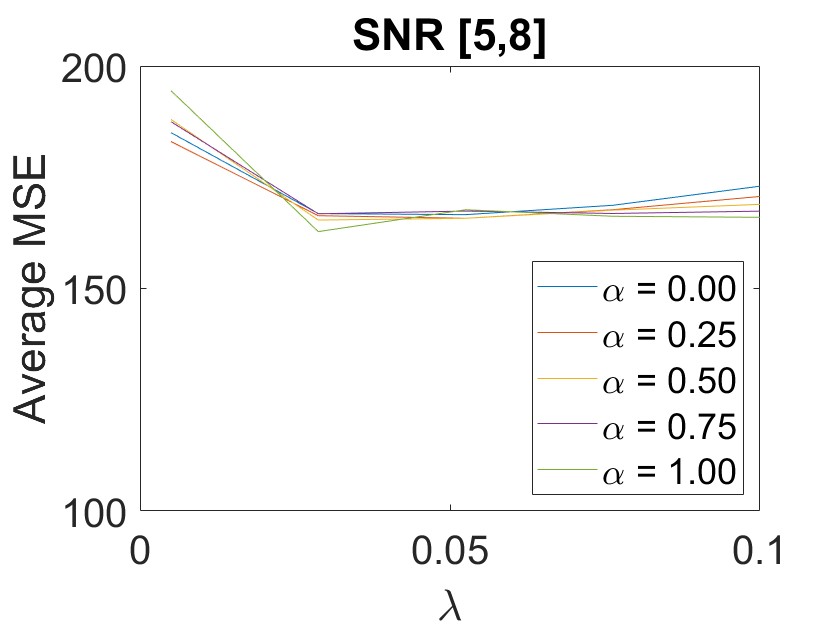}}
               \hfil
      \subfloat[\label{c}]{%
       \includegraphics[width=0.3\textwidth]{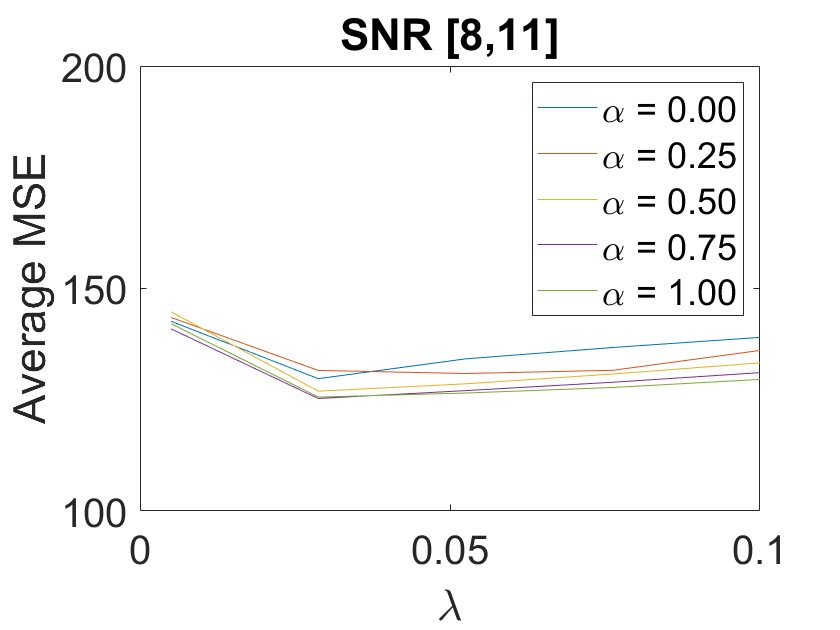}}\\

 \caption{\hl{CV Error for (a) SNR [2,5], (b) SNR [5,8], and (c) SNR [8,11]}}
  \label{fig:CVError} 
\end{figure}

\hl{Following feature extraction, we fitted the MGR-ASGL model, which contains two tuning parameters $\lambda$ and $\alpha$. To select these tuning parameters, we employed 10-fold cross-validation. First, we randomly partitioned the training set into 10 equally sized folds. We note that the indices in these folds were held constant when performing model fitting on each dataset. For this technique, we set aside one fold as a validation set and let the other nine folds comprise the CV-training set. Then we standardized the CA-FPC scores and the $\ln{TTF}$ values in the CV-training set using their respective sample means and sample standard deviations. Using the means and standard deviations of the CA-FPC scores in the CV-training set, we also standardized the CA-FPC scores in the validation set. For each candidate pair $(\lambda,\alpha)$, we fitted the MGR-ASGL model to the CV-training set. The fitted model provides estimates of the regression coefficients, precision parameter, and mixing coefficient for each failure mode. For validation sample $i$, we predicted the standardized $\ln{TTF}$ using the weighed average of the regression predictons for each failure mode as shown in Eq \ref{CVPred},}

\begin{equation}\label{CVPred}
    \hat{y}_i=\sum_{k=1}^2\hat{\pi}_k(\hat{\beta}_{0,k}+\mathbf{x}_i^T\hat{\boldsymbol{{\beta}}_k})
\end{equation}
\hl{where $\hat{\pi}_k$ is the estimated mixing coefficient for failure mode $k$, $\mathbf{x}_i$ is the standardized CA-FPC score vector for validation system $i$, and $(\hat{\beta}_{0,k},\hat{\boldsymbol{{\beta}}_k})=(\hat{\varphi}_{0,k}/\hat{\rho}_k,\hat{\boldsymbol{{\varphi}}_k}/\hat{\rho}_k)$ are the failure mode-specific regression coefficients. We then transformed the prediction back to the $\ln{TTF}$ scale using the mean and standard deviation of the CV-training responses and exponentiated the result to obtain our prediction $\hat{TTF}_i$. For each validation fold, we calculated the mean squared error between the observed and predicted TTF values as shown in Eq \ref{CVError}.}

\begin{equation}\label{CVError}
    MSE_f(\lambda,\alpha)=(1/32)\sum_{i=1}^{32}(TTF_i-\hat{TTF}_i)^2
\end{equation}
\hl{We repeated this process over 5 equally spaced values of $\lambda$ between 0.005 and 0.100 and 5 equally spaced values of $\alpha$ between 0 and 1. We then averaged the MSEs across the 10 folds for each $(\lambda,\alpha)$ pair and selected the pair that minimized the fold-wise averaged MSE, or CV MSE. In Figure \ref{fig:CVError}, we display the averaged cross-validation errors for each SNR scenario.}
\hl{As shown in Table \ref{tab:SimTuningParamSelect}, the minimum error occurs when $\lambda=0.0288$ in each scenario. Furthermore, $\alpha$ equals 1.00 for the first two scenarios and 0.75 for the $[8,11]$ scenario. Despite this consistency in tuning parameter selection, the minimum cross-validation error varies significantly between SNR scenarios. Some potential reasons for this include that the simulation methodology samples an independent SNR value for each trajectory. This means that for a particular sensor, the SNR varies between samples and the relative differences between trajectory SNRs is not maintained across different SNR scenarios. Another reason is that the added noise affects CA-FPCA. Applying CA-FPCA to the second dataset resulted in more features extracted than the other two datasets, which may have raised prediction error on the validation folds.

After determining the optimal tuning parameters, we use the entire training set to fit the MGR-ASGL model. When fitting this model, we standardize both the CA-FPC scores and the $\ln{TTF}$. Column 3 of Table \ref{tab:SimPerformance} displays how accurately the model recovered the latent failure modes and what sensors were selected. For the $[2,5]$ dataset, several of the observations were clustered into one failure mode, which explains why the accuracy for FM 2 is far greater than the accuracy for FM 1. For the two higher SNR datasets, the accuracy sharply improves. This could be due to the reduced noise and the decrease in significance of the weakly informative sensors. This decrease is apparent in the $[8,11]$  dataset, where one sensor is removed for each failure mode.

Column 4 of Table \ref{tab:SimPerformance} displays which sensors were selected for each SNR scenario while Table \ref{tab:l2Sim} lists the sensors in descending order of $\|\hat{\boldsymbol{{\varphi}}_k}\|_2$, the $l_2$-norm of the estimated regression coefficients from sensor $k$. To select sensors, we removed any sensors for which this value was less than $5\times10^{-3}$. In the first two datasets, no sensors were removed. Despite this, Table \ref{tab:l2Sim} shows that the strongly informative sensors are usually represented among the most significant sensors. Sensor 19 is interesting in that it was designed to be strongly informative for both failure modes. However, it only appears in the top 4 for FM 1 when the SNR was lowest. We should note that Sensor 19 placed no worse than 6th on the list of most significant sensors for FM 1. The reasons Sensor 19 may not have been represented in the top 4 are that Sensor 19 decreases for both failure modes so it may not be as useful for distinguishing failure modes and that multiple weakly informative sensors may have had too large correlation with the true degradation rate. Reducing the range of the correlations for the weakly informative sensors may have resulted in more of them not being selected and revealing Sensor 19 to be more informative. As it stands, the model is attempting to identify the best features for predicting $\ln{TTF}$, and the features from weakly informative sensors still have prognostic value. In the next subsection, we utilize the identified latent failure modes and selected sensors to predict RUL.}

\begin{table}[h]
    \centering
    \caption{\hl{Tuning parameter selection for Case Study 1}}
    \setlength{\tabcolsep}{1.2em}
    \begin{tabular}{*4c}
    \toprule
    {\textbf{$SNR_{p}^k$}} & {\textbf{[2,5]}} & {\textbf{[5,8]}} & {\textbf{[8,11]}}\\
    \midrule
    $\lambda$ & 0.0288 & 0.0288 & 0.0288\\
    $\alpha$ & 1 & 1 & 0.75\\
    \bottomrule
\end{tabular}\label{tab:SimTuningParamSelect}
\end{table}

\begin{table}[h]
    \centering
    \caption{\hl{MGR-ASGL performance for case study 1}}
    \setlength{\tabcolsep}{1.2em}
    \begin{tabular}{*5c}
    \toprule
    {\textbf{$SNR_p^k$}} & {\textbf{FM}} & {\textbf{Accuracy}} & {\textbf{Sensors Selected}}\\
    \midrule
    {\textbf{[2, 5]}} & 1 & 0.5500 & 1:20\\
    {} & 2 & 0.9125 & 1:20\\
    \midrule
    {\textbf{[5, 8]}} & 1 & 0.8875 & 1:20\\
    {} & 2 & 0.8313 & 1:20\\
    \midrule
    {\textbf{[8, 11]}} & 1 & 0.8375 & 1,3:20\\
    {} & 2 & 0.8438 & 1:3,5:20\\
    \bottomrule
\end{tabular}\label{tab:SimPerformance}
\end{table}

\subsection{RUL Prediction}

\hl{Each of the datasets consists of 80 test systems (40 from each failure mode). For each test 
system, we observe degradation up until the following life percentiles (in retrospect): 10\%, 
20\%, 30\%, 40\%, 50\%, 60\%, 70\%, 80\%, and 90\%. We smooth the training and the test 
signals using a bandwidth parameter of 0.5. Next, we perform MFPCA, selecting the number of 
components using only the training systems that survived up to the observation time $t^*$ 
corresponding to each life percentile. For a given test signal, we calculate its MFPC scores. 
The active failure mode is diagnosed by being assigned to the most frequent failure mode 
present in its closest $0.1N_{t^*}$ neighbors. Once the active failure mode is identified, 
the MFPC scores of the test signals are updated and recomputed using the subset of training 
signals associated with the active failure mode. Next, we fit the weighted regression model 
and predict the RUL using the scores of the test signal. Prediction accuracy is calculated 
retrospectively. We compute the ``\textit{Estimated Life}" as the current operating time 
plus the predicted RUL. A relative error is then computed using Eq (\ref{RUL}).}
\begin{equation}\label{RUL}
    \text{Relative Error} = \frac{\mid \text{Estimated Life} - \text{Actual Life} \mid}
    {\text{Actual Life}} \times 100\%
\end{equation}
\begin{table}[h]
\centering
\caption{\hl{$l_2$-norms of four most informative sensors for FM1 \& FM2}}
\begin{tabular}{*5c}
\toprule
{\textbf{$SNR_p^k$}} & \multicolumn{2}{c}{\textbf{FM 1}} &  \multicolumn{2}{c}{\textbf{FM 2}}\\
{} & {\textbf{Sensor ID}} & {\textbf{$l_2$-norm}} & {\textbf{Sensor ID}} & {\textbf{$l_2$-norm}} \\
\midrule
{} & 5 & 1.464 & 9 & 1.561 \\
{\textbf{[2, 5]}} &12 & 1.159 & 7 & 1.456 \\
{} &16 & 1.104 & 19 & 0.554 \\
{} &19 & 0.723 & 6 & 0.373 \\
\midrule
{} &12 & 1.886 & 9 & 2.623 \\
{\textbf{[5, 8]}} &5 & 1.732 & 7 & 0.854 \\
{} &16 & 0.949 & 3 & 0.554 \\
{} &3 & 0.348 & 19 & 0.409 \\
\midrule
{} &12 & 1.846 & 9 & 2.669 \\
{\textbf{[8, 11]}} &5 & 1.502 & 7 & 0.661 \\
{} &16 & 0.758 & 17 & 0.400 \\
{} &18 & 0.430 & 19 & 0.398 \\
\bottomrule
\end{tabular}\label{tab:l2Sim} 
\end{table}

\begin{figure}[h]
    \centering
    \includegraphics[width=\columnwidth]{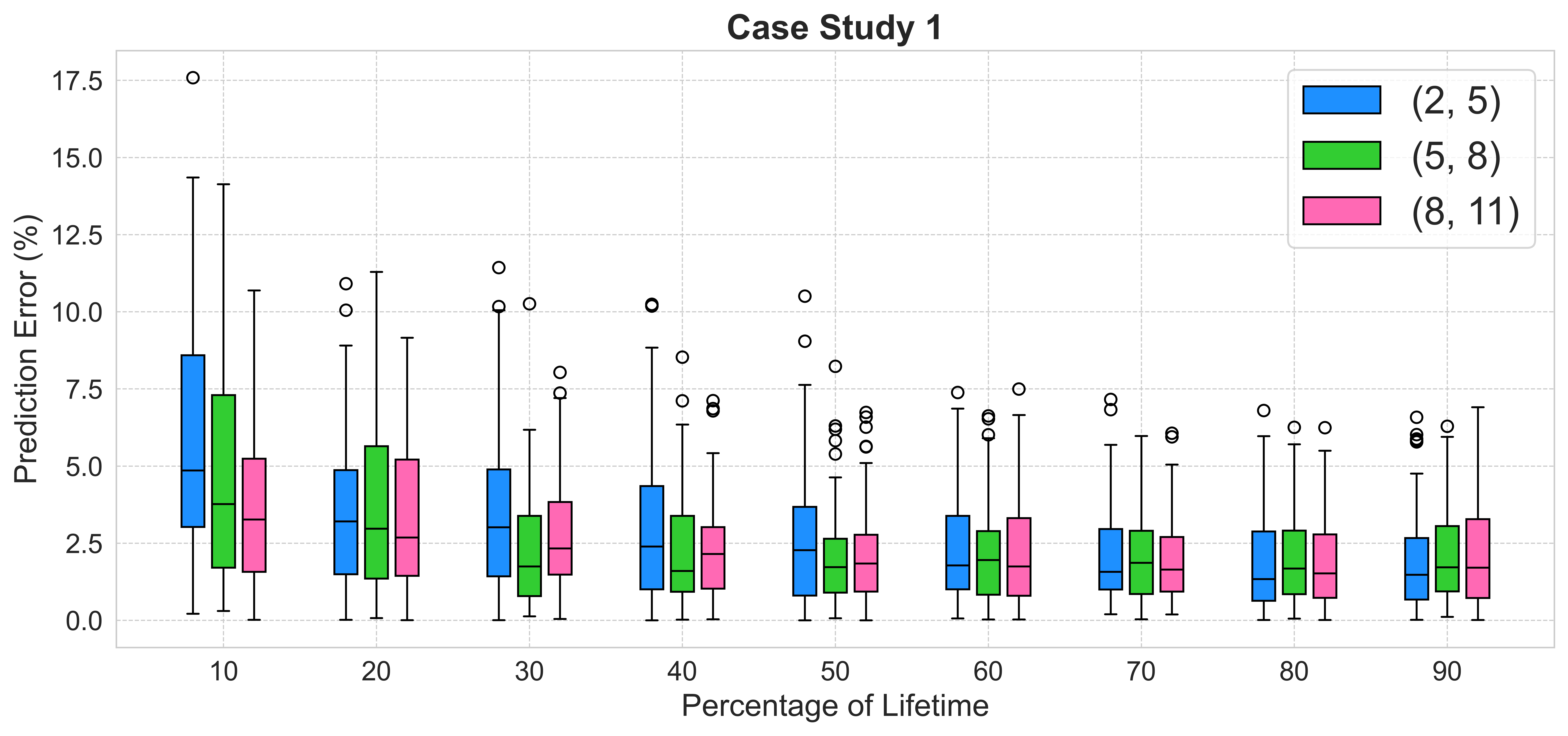}
    \caption{\hl{Prediction error (in $\%$) for all test systems of case study 1 under three different SNRs, i.e., [2, 5], [5, 8], and [8, 11]}}
    \label{fig:SimRUL}
\end{figure}

\hl{The performance of our methodology on predicting RUL is shown in Figure \ref{fig:SimRUL}. Figure \ref{fig:SimRUL} shows that as the percentage of system TTF increased from 10$\%$ to 50$\%$, the median and the interquartile range of the relative errors generally decreased. This indicates that the accuracy and precision of RUL predictions tend to improve as the system degrades. Comparing the performance of our methodology on different SNR scenarios, we see that for this percentage range, both the $[5,8]$ and $[8,11]$ scenarios had more accurate and precise RUL predictions than the $[2,5]$ scenario. An exception occurs at 20$\%$, where the interquartile range is smaller for the $[2,5]$ dataset than for the other two SNR scenarios. Interestingly, between 30$\%$ and 50$\%$, the median of the relative error for the $[5,8]$ scenario was less than that for the $[8,11]$ scenario. However, the opposite was true regarding the interquartile range. This result may have been caused by our sensor selection methodology removing one sensor from each failure mode. Retaining an extra sensor may have resulted in higher accuracy for the $[5,8]$ scenario, but also higher variance.
Following 50$\%$, the median and interquartile range ceased to significantly improve for both increasing percentage of system TTF and increasing SNR. A potential cause for this result is that the significance of sensors may depend on how long the system has survived. In our methodology, we select sensors based on features extracted at the time of the earliest recorded failure so as to not remove observed failures from the dataset, which could affect the performance of the mixture model. However, the cost is that the sensors selected may be suboptimal at later stages of system degradation. Another potential reason is that the upper bound on the distribution of the correlation coefficient for weakly significant sensors may be too large such that too many sensors are retained.}

%% file: CaseStudy2.tex
\begin{figure}[h]
    \centering
        \includegraphics[width=0.75\columnwidth]{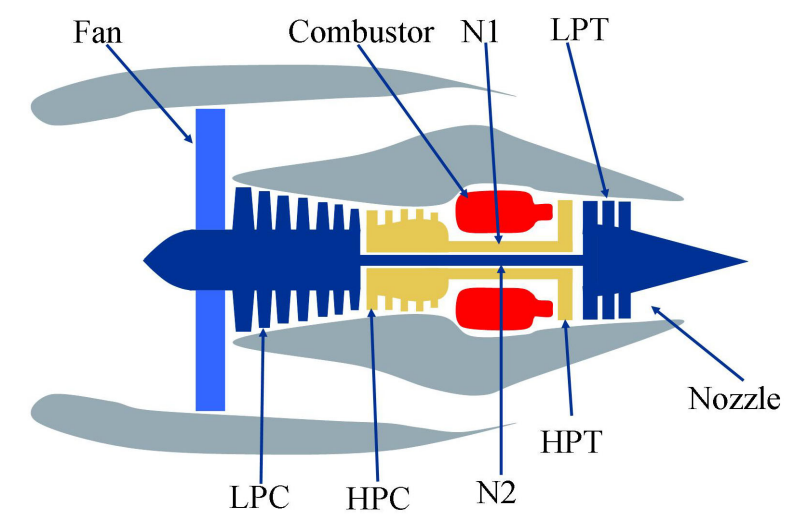}
  \caption{\hl{The system diagram used in Case Study 2, adapted from \cite{frederick2007user}}}
  \label{fig:CMAPSS}
\end{figure}

\hl{To further evaluate the proposed methodology on a realistic prognostics benchmark, we apply it to NASA’s dataset called the \textit{Commercial Modular Aero-Propulsion System Simulation}, or C-MAPSS}$\footnote{\hl{The C-MAPSS dataset is publicly available at} 
\url{https://data.nasa.gov/dataset/cmapss-jet-engine-simulated-data}. \hl{This dataset has four different subsets: FD001, FD002, FD003, and FD004}}$. \hl{The dataset contains multivariate sensor measurements from turbofan engines operating from a good-as-new state to system failure, reflecting the run-to-failure monitoring paradigm that is expected in a DSH subsystem. A diagram of one of the engines is provided in Figure \ref{fig:CMAPSS}. In this dataset, engine degradation is driven by two fault modes: fan degradation and high-pressure compressor (HPC) degradation. These degradation processes are monitored through a set of onboard sensors that record temperature, pressure, and flow-related measurements throughout the engine’s operating life. Further details regarding the C-MAPSS dataset and its simulation setup are provided in \cite{frederick2007user, Saxena2008}.}

\begin{figure}
    \centering
        \includegraphics[width=\columnwidth]{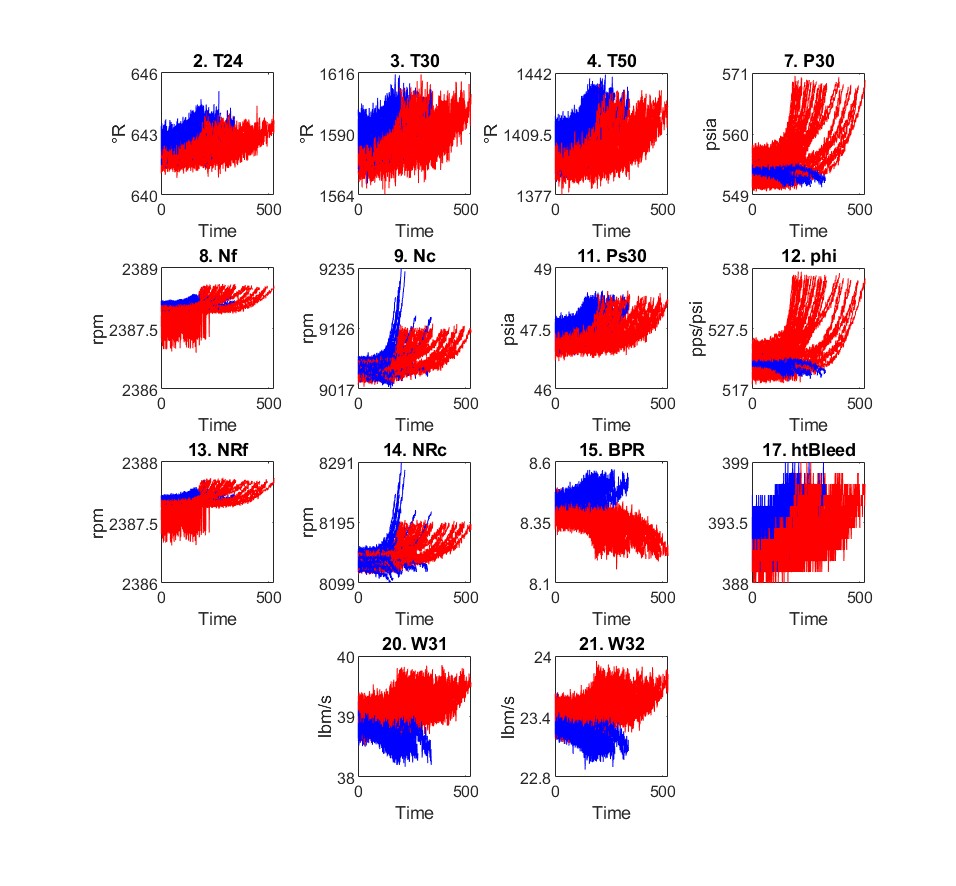}
  \caption{\hl{Sensor data used for the offline step in case study 2. HPC failure (blue), Fan failures (red)}}
  \label{fig:realSensorData}
\end{figure}

\hl{In this paper, we utilize the FD003 subset of the C-MAPSS data, which comprises a training set and a test set. In the training set, the sensors generate trajectories from 21 sensors for 100 engines as they degrade from a good-as-new state to failure. The training set does not provide the true failure mode labels, mirroring the unlabeled failure mode setting of a DSH. %where post-failure expert diagnosis is constrained by communication delays. 
However, they can be inferred from the fact that the FD001 subset consists of only HPC failures, while the FD003 subset consists of both HPC and fan failures. The training set of FD003 is displayed in Figure \ref{fig:realSensorData}, and descriptions of the sensors are provided in Table~\ref{tab:CMAPPS_sensors}. Sensors 1, 5, 6, 10, 16, 18, and 19 contain little to no information and were therefore removed from the analysis. In Figure \ref{fig:realSensorData}, HPC failures are displayed in blue and fan failures are displayed in red. For each failure mode, C-MAPSS models degradation through progressive changes in component health parameters, including efficiency, flow capacity, and pressure ratio, which propagate through the coupled engine dynamics and produce distinct multivariate sensor trajectories \cite{frederick2007user}. In particular, HPC degradation directly affects the compression characteristics of the engine core, whereas fan degradation primarily alters the upstream fan/bypass-flow characteristics. Consequently, the two failure modes exhibit distinct degradation patterns across several sensors. In particular, FM1 (HPC failure, blue) is characterized by an increasing trend in the BPR sensor and decreasing trends in the P30, phi, W31, and W32 sensors, whereas FM2 (fan failure, red) generally exhibits trends in the opposite direction for these sensors.} %Five sensors (P30, phi, BPR, W31, and W32) exhibit signals with trends moving opposite to the HPC failures, and thus we infer these are fan failures. 
%Notably, P30 refers to pressure recorded at the HPC outlet. The association between HPC degradation and a decrease in pressure is straightforward, considering the role of the HPC is to raise the pressure of the airflow. However, pressure increases associated with fan degradation is the result of actions taken by the fan-speed controller to maintain demanded fan speed, which result in new operating points for the engine. 
While this figure displays the true failure modes, we emphasize that our methodology considers that these failure mode labels are unknown.
The testing set is similar to the training set in that it too consists of 100 engines monitored by 21 sensors. However, the degradation for each engine in the test set is only partially observed. Thus, our objective is to predict the RUL of these engines under unknown failure mode labels.

%The 21 sensors monitor the systems from a good-as-new state to engine failure, reflecting the run-to-failure monitoring paradigm that is expected in autonomous DSH operations. However, sensors 1, 5, 6, 10, 16, 18, and 19 contain little to no information and are thus removed from the analysis. While the dataset provides the potential causes of failure, it does not provide the failure labels, mirroring the unlabeled failure setting of a DSH where post-failure expert diagnosis is constrained by communication delays. Furthermore, the test set consists of systems for which degradation is only partially observed, together with their corresponding failure times. Therefore, our goal is to use our methodology to predict the remaining life of each testing unit following the final cycle at which their degradation was observed.

Our first step is to perform traditional FPCA on the signals for each sensor. We obtain 
initial failure mode labels for the training set by applying K-means clustering to all 
FPC scores. Then, for each sensor, we apply K-means clustering to the FPC scores associated 
with that sensor to obtain the covariate information needed to perform CA-FPCA. Following 
feature extraction, we use 3-fold cross-validation to determine $(\lambda, \alpha)$. The 
range of values for these tuning parameters is identical to that of the simulation case 
study. The combination of tuning parameters that minimizes the mean squared error across 
the three folds is $(\lambda_{MSE}, \alpha_{MSE}) = (0.1089, 1)$. Table 
\ref{tab:CMAPPS_sensorselection} displays the $l_2$-norms of the regression coefficients 
for each sensor. The sensors selected are those with positive $l_2$-norms.
\begin{table}[h]
\centering
\caption{Sensors used in case study 2 (C-MAPPS\cite{Saxena2008})}
\begin{tabular}{*3c}
\toprule
{\textbf{Sensor ID}} & {\textbf{Sensor Name}} & {\textbf{Functionality}}\\
\midrule
1 & T2 & Total temperature at fan inlet \\
2 & T24 & Total temperature at LPC outlet\\
3 & T30 & Total temperature at HPC outlet\\
4 & T50 & Total temperature at LPT outlet\\
5 & P2 & Pressure at fan inlet\\
6 & P15 & Total pressure in bypass-duct\\
7 & P30 & Total pressure at HPC outlet\\
8 & Nf & Physical fan speed\\
9 & Nc & Physical core speed\\
10 & epr & Engine pressure ratio (P50/P2)\\
11 & Ps30 & Static pressure at HPC outlet\\
12 & phi & Ratio of fuel flow to Ps30\\
13 & NRf & Corrected fan speed\\
14 & Nrc & Corrected core speed\\
15 & BPR & Bypass ratio\\
16 & farB & Burner fuel-air ratio\\
17 & htBleed & Bleed enthalpy\\
18 & Nfdmd & Demanded fan speed\\
19 & PCNfRdmd & Demanded corrected fan speed\\
20 & W31 & HPT coolant bleed\\
21 & W32 & LPT coolant bleed\\
\bottomrule
\end{tabular}\label{tab:CMAPPS_sensors} 
\end{table}

In addition to performing sensor selection, the MGR-ASGL model provides cluster labels for 
each observed failure. These labels are utilized for the online prediction portion of the 
methodology. For the online prediction step, we first smooth the degradation signals using 
5-fold cross-validation to select the bandwidth parameter. After smoothing, we perform MFPCA 
on signals from the sensors selected across both failure modes to build our classifier. Then, 
we classify the failure mode based on the $0.1N_{t^*}$ nearest neighbors. Given this 
classification, we perform MFPCA on signals from the sensors selected for the classified 
failure mode. Finally, we perform Weighted Time-varying Functional Regression to regress 
$\ln TTF$ on the MFPC scores using leave-one-out cross-validation to select the tuning 
parameter.

Table \ref{tab:CMAPPS_sensorselection} displays the $l_2$-norms of the regression 
coefficients for each sensor resulting from the offline MGR-ASGL step. A positive $l_2$-norm 
indicates that the sensor was selected as informative for that failure mode. Notably, the 
selected sensors differ between FM 1 and FM 2, confirming that the MGR-ASGL model 
successfully identifies mode-specific sensor subsets without access to failure mode labels. 
This is a particularly desirable property for DSH health monitoring, where the active failure 
mode is unknown, and sensor relevance is expected to vary across subsystem degradation 
mechanisms.

\subsection{Comparison to Baselines}

To analyze our methodology, we compare its performance to that of \cite{Chehade2018} 
(hereafter referred to as ``\textbf{Chehade et al. (2018)}''), whose methodology assumes knowledge of the failure modes but does 
not perform sensor selection outside of removing the uninformative sensors, and \cite{Wu2023} 
(hereafter referred to as ``\textbf{Wu and Li (2023)}''), whose methodology assumes knowledge of a fraction of the training set and 
does perform sensor selection. Both baselines, therefore, have access to information that would 
be unavailable in an autonomous DSH setting, namely, labeled failure modes or partially labeled 
training data, making this comparison a conservative test of our methodology.
% \begin{table}[h]
% \centering
% \caption{Sensor Selection for Case Study 2}
% % \setlength{\tabcolsep}{1.2em}
% \begin{tabular}{*3c}
% {\textbf{Sensor ID}} & {\text{FM 1}} & {\text{FM2}}\\
% \hline
% \textbf{2} & 0.266 & 0 \\ 
% \textbf{3} & 0.584 & 0 \\ 
% \textbf{4} & 0.897 & 0.033 \\ 
% \textbf{7} & 0 & 0.877 \\ 
% \textbf{8} & 0.117 & 0 \\ 
% \textbf{9} & 0.545 & 0.291 \\
% \textbf{11} & 0 & 0 \\
% \textbf{12} & 0 & 1.148\\
% \textbf{13} & 0.90 & 0\\
% \textbf{14} & 0.148 & 0.119\\
% \textbf{15} & 0.501 & 0\\
% \textbf{17} & 0.230 & 0.042\\
% \textbf{20} & 0 & 0.573\\
% \textbf{21} & 0.116 & 0.318\\
% \hline
% \end{tabular}\label{tab:CMAPPS_sensorselection}
% \end{table}
\begin{table}[h]
\centering
\caption{Sensor Selection for Case Study 2}
\setlength{\tabcolsep}{0.45em}
\begin{tabular}{c|*{14}{c}}
\textbf{Sensor ID} 
& \textbf{2} & \textbf{3} & \textbf{4} & \textbf{7} 
& \textbf{8} & \textbf{9} & \textbf{11} & \textbf{12} 
& \textbf{13} & \textbf{14} & \textbf{15} & \textbf{17} 
& \textbf{20} & \textbf{21} \\
\hline
\textbf{FM 1} 
& 0.266 & 0.584 & 0.897 & 0 
& 0.117 & 0.545 & 0 & 0 
& 0.90 & 0.148 & 0.501 & 0.230 
& 0 & 0.116 \\
\textbf{FM 2} 
& 0 & 0 & 0.033 & 0.877 
& 0 & 0.291 & 0 & 1.148 
& 0 & 0.119 & 0 & 0.042 
& 0.573 & 0.318 \\
\hline
\end{tabular}
\label{tab:CMAPPS_sensorselection}
\end{table}

\begin{figure}[H]
    \centering
        \includegraphics[width=0.95\columnwidth]{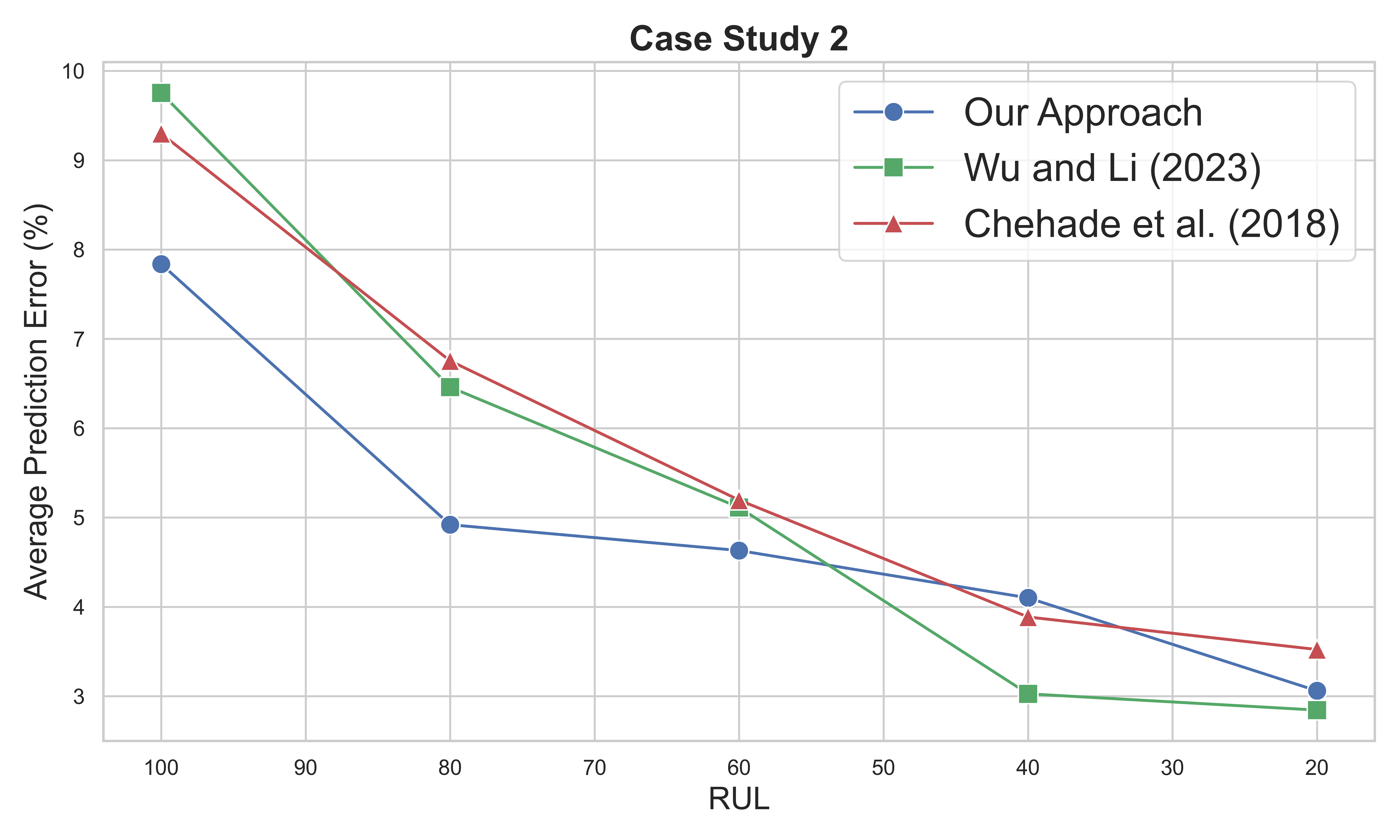}
  \caption{\hl{Average Prediction error ($\%$) in case study 2}}
  \label{fig:avgpredRUL_real} 
\end{figure}

Figure \ref{fig:avgpredRUL_real} displays the relative prediction error averaged over all 
test systems with RUL at most 100, 80, 60, 40, and 20 cycles. From this figure, our 
methodology becomes more accurate as the RUL decreases, which is the operationally most 
critical regime for a DSH where late-life maintenance decisions must be made autonomously. Furthermore, our methodology is more accurate than both baselines 
when making longer predictions. For test units whose RUL is at most 20, our accuracy is 
between that of \textbf{Chehade et al. (2018)} and \textbf{Wu and Li (2023)}. Only at RUL at most 40 does our 
methodology perform worse than both baselines.
\begin{figure}[h]
    \centering
        \includegraphics[width=0.95\columnwidth]{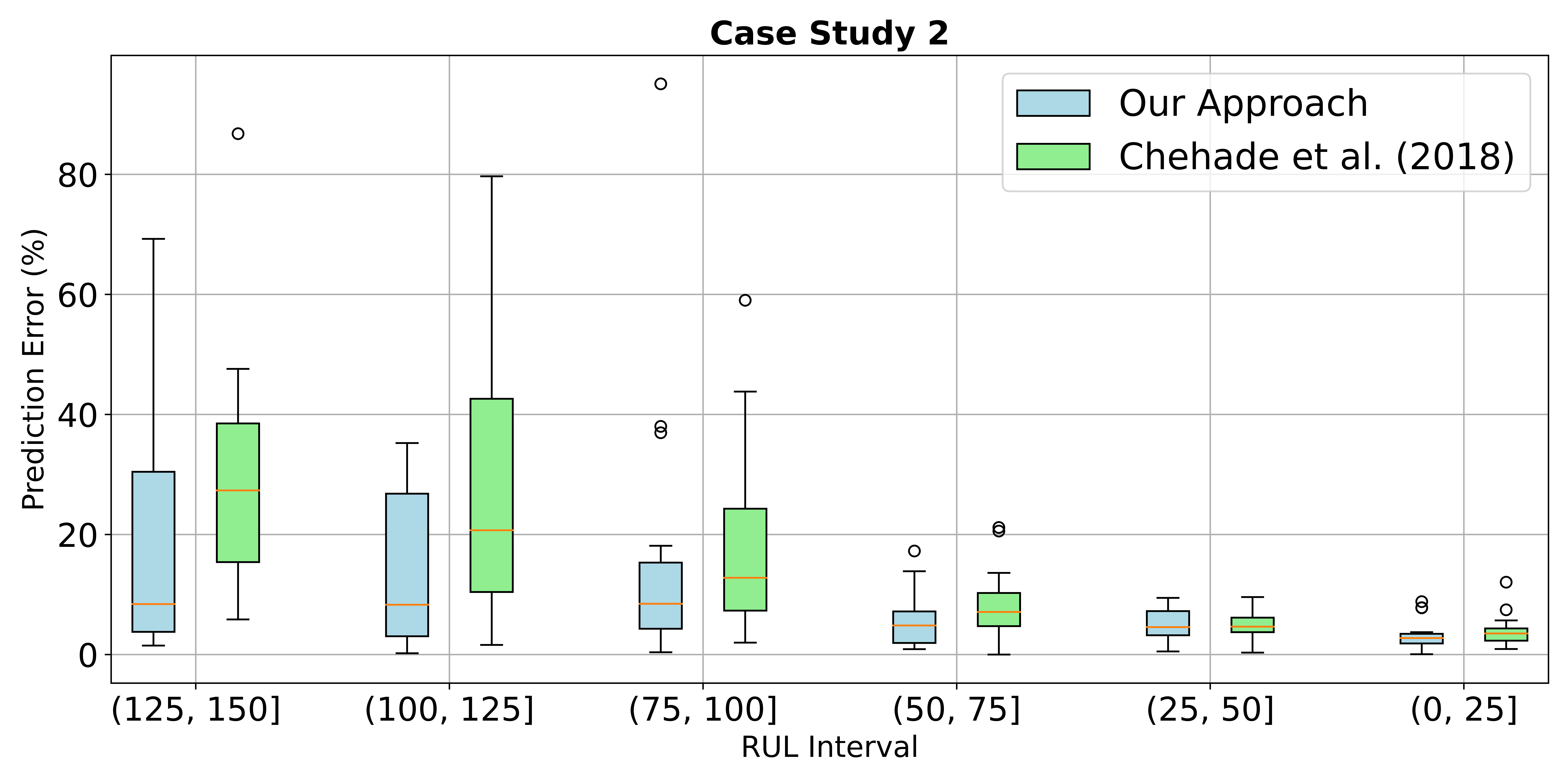}
  \caption{\hl{Prediction error ($\%$) for different RUL intervals in case study 2}}
  \label{fig:realRUL} 
\end{figure}

To further analyze our methodology, Figure \ref{fig:realRUL} displays boxplots showing 
the distribution of the relative prediction error for test units with remaining life in 
the following intervals: $[0,25]$, $[26,50]$, $[51,75]$, $[76,100]$, $[101,125]$, 
$[126,150]$. The results indicate that our methodology achieves a lower median relative 
error than \textbf{Chehade et al. (2018)} in all intervals while maintaining comparable variability. 
However, the variability for systems with RUL between 25 and 50 is higher for our 
methodology than for \textbf{Chehade et al. (2018)}, indicating a targeted area for future improvement. 
The general trend is that sensor selection resulted in more accurate RUL prediction than 
no sensor selection. Furthermore, we were capable of achieving more accurate RUL prediction 
than both baselines despite having no information about the actual failure mode, which is 
the operationally realistic scenario for an autonomous DSH.

%% file: Conclusion.tex
\hl{This paper presented a two-phase framework for predicting RUL in DSH systems. Our framework is beneficial for DSHs because it does not rely on labeled failure modes, which may not be readily available. Instead, the first phase utilizes telemetry data from multiple sensors to jointly identify the latent failure modes and select the optimal subset of sensors for each failure mode. For this task, we proposed our MGR-ASGL model. While this phase can be performed either on Earth or onboard the DSH, we expect the second phase to be performed onboard the DSH exclusively. The second phase consists of using the sensors selected in Phase 1 to diagnose the active failure mode of a degrading asset and then predict its RUL. This task is performed adaptively such that whenever new data is available, model parameters are refit so that diagnosis and prognostics can be updated in real-time. Furthermore, this does not require a human operator. Thus, the second phase of the framework is conducive to self-reliance of a DSH.

In this paper, we evaluated our methodology using two case studies. In the first case study, we showed that our methodology could accurately identify the latent failure modes with over 80\% accuracy when the SNR was both moderate and high. While our methodology did not remove sensors, it did show that the sensors designed to be highly informative were usually the most significant sensors. In the second case study, we were able to achieve a sparse sensor selection and outperform two comparable methodologies with regard to predictions made early in system life. This could be valuable in a DSH, where lead times for spare parts provision are expected to be substantial due to the DSH's distance from Earth.

One limitation of our methodology is that we do not consider different operating regimes. Therefore, future work can consist of incorporating time-varying operating states into the model. Our analysis also showed room for improvement for predictions made closer to a system's TTF. Therefore, further future work could include relaxing the assumption that the number of failure modes are known, synthesizing the two phases of our framework, and adapting the mixture model for nonlinear regression. Assuming that the number of failure modes are unknown promotes further self-reliance since it removes an element of human intervention. Synthesizing the two phases increases the efficiency of the framework. Furthermore, this may also enable a sensor selection that adapts to the degradation state of the asset. Finally, a mixture of nonlinear regression could both improve latent failure mode identification and reduce the number of sensors required for RUL prediction.}

%% file: Appendix.tex
\section{Log-Likelihood}
\label{app1}
\begin{definition}\label{def1}
    The log-likelihood function for the  MGR model is given by equation (\ref{eq12}). This function is called as the "\textit{incomplete-data log-likelihood}" (IDLL), since we do not know the cause of the observed failure in the data. 
    \begin{equation}\label{eq12}
    \ell(\boldsymbol{\Theta};\boldsymbol{Y})=\sum_{i=1}^N\ln\sum_{k=1}^K\pi_k\frac{\rho_k}{\sqrt{2\pi}}\exp (-\frac{1}{2}(\rho_k y_i-\psi_{0, k}-\sum_{p=1}^P \boldsymbol{x}_{i, p}^T \boldsymbol{\psi}_{p, k})^2) 
    \end{equation}
\end{definition}

\begin{definition}\label{def2}
   The ``\textit{complete-data log-likelihood}" (CDLL) given by equation (\ref{eq13}) assumes knowledge of the underlying causes of the observed failure in the data. 
    \begin{equation}\label{eq13}
        \ell_C(\boldsymbol{\Theta} ; \boldsymbol{Y}, \mathbb{Z})=\sum_{i=1}^N \sum_{k=1}^K \boldsymbol{Z}_i[k] \ln \pi_k\left(\frac{\rho_k}{\sqrt{2 \pi}}\right) e^{\left(-\frac{1}{2}\left(\rho_k y_i-\psi_{0, k}-\sum_{p=1}^P \boldsymbol{x}_{i, p}^T \boldsymbol{\psi}_{p, k}\right)^2\right)}
    \end{equation}
\end{definition}

The IDLL given in equation (\ref{eq12}) is difficult to optimize globally \cite{murphy2012}. Instead, we search for a local optimum by using EM algorithm, an iterative algorithm suitable for fitting probability models with latent variables. To derive the steps of the EM algorithm, we note that since $\ln(.)$ is a concave function We use Jensen's inequality to form a lower bound on the IDLL. This lower bound is the expectation of the CDLL given in equation (\ref{eq13}) and shown in Lemma \ref{lemma1}.

\begin{lemma}\label{lemma1}
    The expectation of the CDLL w.r.t distribution $g$ is a lower bound for the IDLL i.e., \begin{equation*}\mathbb{E}_{\boldsymbol{g}}\left[\ell_C(\boldsymbol{\Theta}; \boldsymbol{Y}, \mathbb{Z}) \right] \leq \ell(\boldsymbol{\Theta}; \boldsymbol{Y})
   \end{equation*}
\end{lemma}

\begin{proof}
    Let $g$ be an arbitrary distribution on $\boldsymbol{Z}_i[k]$. Then we can re-write equation (\ref{eq12}) as follows: 
    \begin{equation*}
    \ell(\boldsymbol{\Theta};\boldsymbol{Y})=\sum_{i=1}^N\ln\left(\sum_{k=1}^K\frac{g\left(\boldsymbol{Z}_i[k]\right)p\left(y_i,\boldsymbol{Z}_i[k];\boldsymbol{\Theta}\right)}{g\left(\boldsymbol{Z}_i[k]\right)}\right)\vspace*{-1em}
    \end{equation*}
    where, 
    \begin{equation*}
        p\left(y_i,\boldsymbol{Z}_i[k]; \boldsymbol{\Theta}\right)=\pi_k \cdot \left(\frac{\rho_k}{\sqrt{2\pi}}e^{-\frac{1}{2}\left(\rho_k y_i-\psi_{0, k}-\sum_{p=1}^P \boldsymbol{x}_{i, p}^T \boldsymbol{\psi}_{p, k}\right)^2}\right)
    \end{equation*}
    
    Since $\ln(.)$ is concave, from Jensen's inequality we can write that, 
    \begin{equation*}
        \ell(\boldsymbol{\Theta};\boldsymbol{Y})\geq\sum_{i=1}^N\sum_{k=1}^Kg\left(\boldsymbol{Z}_i[k]\right)\ln{\frac{p\left(y_i,\boldsymbol{Z}_i[k];\boldsymbol{\Theta}\right)}{g\left(\boldsymbol{Z}_i[k]\right)}}
    \end{equation*}
and thus, 
    \begin{equation*}
         \ell(\boldsymbol{\Theta};\boldsymbol{Y})\geq\mathbb{E}_{\boldsymbol{g}}\left[\ell_C(\boldsymbol{\Theta}; \boldsymbol{Y}, \mathbb{Z}) \right]-\sum_{i=1}^N\sum_{k=1}^Kg\left(\boldsymbol{Z}_i[k]\right)\ln{g\left(\boldsymbol{Z}_i[k]\right)}
   \end{equation*}
 The second term on the right side of the inequality is the negative entropy of the $g$ distribution. Since it is positive, the inequality holds. 
\end{proof}

%% file: main.bbl
\begin{thebibliography}{72}
\newcommand{\enquote}[1]{``#1''}
\providecommand{\natexlab}[1]{#1}
\providecommand{\url}[1]{\texttt{#1}}
\providecommand{\urlprefix}{URL }
\expandafter\ifx\csname urlstyle\endcsname\relax
  \providecommand{\doi}[1]{\discretionary{}{}{}https://doi.org/#1}\else
  \providecommand{\doi}[1]{\discretionary{}{}{}\urlstyle{rm}\url{https://doi.org/#1}}\fi

\bibitem[{Lane et~al.(2025)Lane, Nilvarna, Abbey~Jr, and Kezirian}]{lane2025probabilistic}
Lane, K., Nilvarna, A., Abbey~Jr, G.~W., and Kezirian, M.~T., \enquote{Probabilistic risk assessment (Pra) for artemis and future space exploration programs,} \emph{Acta Astronautica}, 2025.
\newblock \doi{10.1016/j.actaastro.2025.06.045}.

\bibitem[{Creech et~al.(2022)Creech, Guidi, and Elburn}]{creech2022artemis}
Creech, S., Guidi, J., and Elburn, D., \enquote{Artemis: An overview of NASA's activities to return humans to the moon,} \emph{2022 ieee aerospace conference (aero)}, IEEE, 2022, pp. 1--7.

\bibitem[{Kirshner and Valerdi(2022)}]{kirshner2022integrating}
Kirshner, M., and Valerdi, R., \enquote{Integrating model-based systems and digital engineering for crewed mars mission planning,} \emph{Journal of Aerospace Information Systems}, Vol.~19, No.~10, 2022, pp. 668--676.
\newblock \doi{10.2514/1.I010986}.

\bibitem[{Fuller et~al.(2022)Fuller, Lehnhardt, Zaid, and Halloran}]{fuller2022gateway}
Fuller, S., Lehnhardt, E., Zaid, C., and Halloran, K., \enquote{Gateway program status and overview,} \emph{Journal of Space Safety Engineering}, Vol.~9, No.~4, 2022, pp. 625--628.
\newblock \doi{10.1016/j.jsse.2022.07.008}.

\bibitem[{Hollaway et~al.(2020)Hollaway, Taylor, and Badger}]{hollaway2020eyes}
Hollaway, D., Taylor, E., and Badger, J., \enquote{When the eyes don’t have it: Autonomous control of deep space vehicles for human spaceflight,} \emph{ASCEND 2020}, 2020, p. 4163.

\bibitem[{Wu(2023)}]{wu2023international}
Wu, X., \enquote{The international lunar research station: China's new era of space cooperation and its new role in the space legal order,} \emph{Space Policy}, Vol.~65, 2023, p. 101537.
\newblock \doi{10.1016/j.spacepol.2022.101537}.

\bibitem[{Gratius et~al.(2024)Gratius, Wang, Hwang, Hou, Rollock, George, Berg{\'e}s, and Akinci}]{gratius2024digital}
Gratius, N., Wang, Z., Hwang, M.~Y., Hou, Y., Rollock, A., George, C., Berg{\'e}s, M., and Akinci, B., \enquote{Digital twin technologies for autonomous environmental control and life support systems,} \emph{Journal of Aerospace Information Systems}, Vol.~21, No.~4, 2024, pp. 332--347.
\newblock \doi{10.2514/1.I011320}.

\bibitem[{Carbone and Loparo(2023)}]{carbone2023fault}
Carbone, M.~A., and Loparo, K.~A., \enquote{Fault detection and diagnosis in spacecraft electrical power systems,} \emph{Journal of Aerospace Information Systems}, Vol.~20, No.~6, 2023, pp. 308--318.
\newblock \doi{10.2514/1.I011136}.

\bibitem[{des Bordes et~al.(2021)des Bordes, Dischinger, Andrews, Allen, and Brown}]{des2021deep}
des Bordes, E., Dischinger, C., Andrews, T., Allen, B., and Brown, M.~D., \enquote{Deep Space Systems Health Management,} \emph{Marshall Space Flight Center Faculty Fellowship Program}, 2021, p.~29.

\bibitem[{Zaccarine and Klaus(2024)}]{zaccarine2024monitoring}
Zaccarine, S.~A., and Klaus, D.~M., \enquote{Monitoring, maintenance and fault management considerations for self-sufficient deep-space habitat operations,} \emph{Acta Astronautica}, Vol. 225, 2024, pp. 376--389.
\newblock \doi{10.1016/j.actaastro.2024.07.055}.

\bibitem[{Yin et~al.(2024)Yin, Zhang, Wang, and Jiang}]{yin2024time}
Yin, M., Zhang, R., Wang, W., and Jiang, J., \enquote{Time-Varying Operating Condition--Based Prediction of the Remaining Useful Life for Aeroengine,} \emph{Journal of Aerospace Information Systems}, Vol.~21, No.~7, 2024, pp. 598--604.
\newblock \doi{10.2514/1.I011223}.

\bibitem[{Rozas et~al.(2025)Rozas, Xie, Gebraeel, and Robinson}]{rozas2025data}
Rozas, H., Xie, W., Gebraeel, N., and Robinson, S., \enquote{Data-driven joint optimization of maintenance and spare parts provisioning: A distributionally robust approach,} \emph{European Journal of Operational Research}, 2025.
\newblock \doi{10.1016/j.ejor.2025.06.025}.

\bibitem[{Guo et~al.(2020)Guo, Li, and Lv}]{guo2020nonlinear}
Guo, R., Li, Y., and Lv, M., \enquote{Nonlinear predictive filter based fault diagnosis of oxygen generation system by using electrolytic water in space station,} \emph{Acta Astronautica}, Vol. 168, 2020, pp. 230--241.
\newblock \doi{10.1016/j.actaastro.2019.12.016}.

\bibitem[{Xiong et~al.(2023)Xiong, Zhou, Ma, Zhang, and Lin}]{Xiong2023}
Xiong, J., Zhou, J., Ma, Y., Zhang, F., and Lin, C., \enquote{Adaptive deep learning-based remaining useful life prediction framework for systems with multiple failure patterns,} \emph{Reliability Engineering \& System Safety}, Vol. 235, 2023, p. 109244.
\newblock \doi{10.1016/j.ress.2023.109244}.

\bibitem[{Cofre-Martel et~al.(2021)Cofre-Martel, Lopez~Droguett, and Modarres}]{cofre2021big}
Cofre-Martel, S., Lopez~Droguett, E., and Modarres, M., \enquote{Big machinery data preprocessing methodology for data-driven models in prognostics and health management,} \emph{Sensors}, Vol.~21, No.~20, 2021, p. 6841.
\newblock \doi{10.3390/s21206841}.

\bibitem[{Oeftering et~al.(2011)Oeftering, Bradish, Juergens, and Lewis}]{oeftering2011component}
Oeftering, R.~C., Bradish, M.~A., Juergens, J.~R., and Lewis, M.~J., \enquote{Component-level electronic-assembly repair (CLEAR) analysis of the problem reporting and corrective action (PRACA) database of the International Space Station on-orbit electrical systems,} Tech. rep., 2011.

\bibitem[{Howe et~al.(2013)Howe, Kennedy, Gill, Smith, and George}]{howe2013nasa}
Howe, S.~A., Kennedy, K.~J., Gill, T.~R., Smith, R.~W., and George, P., \enquote{NASA habitat demonstration unit (HDU) deep space habitat analog,} \emph{AIAA Space 2013 conference and Exposition}, 2013, p. 5436.

\bibitem[{Iverson et~al.(2012)Iverson, Martin, Schwabacher, Spirkovska, Taylor, Mackey, Castle, and Baskaran}]{iverson2012general}
Iverson, D.~L., Martin, R., Schwabacher, M., Spirkovska, L., Taylor, W., Mackey, R., Castle, J.~P., and Baskaran, V., \enquote{General purpose data-driven monitoring for space operations,} \emph{Journal of Aerospace Computing, Information, and Communication}, Vol.~9, No.~2, 2012, pp. 26--44.
\newblock \doi{10.2514/1.54964}.

\bibitem[{Rollock and Klaus(2022)}]{rollock2022defining}
Rollock, A.~E., and Klaus, D.~M., \enquote{Defining and characterizing self-awareness and self-sufficiency for deep space habitats,} \emph{Acta Astronautica}, Vol. 198, 2022, pp. 366--375.
\newblock \doi{10.1016/j.actaastro.2022.06.002}.

\bibitem[{Fang et~al.(2017)Fang, Paynabar, and Gebraeel}]{Fang2017}
Fang, X., Paynabar, K., and Gebraeel, N., \enquote{Multistream sensor fusion-based prognostics model for systems with single failure modes,} \emph{Reliability Engineering \& System Safety}, Vol. 159, 2017, p. 322–331.
\newblock \doi{10.1016/j.ress.2016.11.008}.

\bibitem[{Mirfarah et~al.(2025)Mirfarah, Lund, and Dyke}]{mirfarah2025estimation}
Mirfarah, M., Lund, A., and Dyke, S.~J., \enquote{Estimation of Pressure Leakage Severity for Space Habitats Using Extended Kalman Filter,} \emph{AIAA Journal}, Vol.~63, No.~4, 2025, pp. 1615--1628.
\newblock \doi{10.2514/1.J063825}.

\bibitem[{Hassani.N et~al.(2019)Hassani.N, Jin, and Ni}]{jin2019physics}
Hassani.N, S.~M., Jin, X., and Ni, J., \enquote{Physics-based Gaussian process for the health monitoring for a rolling bearing,} \emph{Acta Astronautica}, Vol. 154, 2019, pp. 133--139.
\newblock \doi{10.1016/j.actaastro.2018.10.029}.

\bibitem[{Kordestani et~al.(2023)Kordestani, Orchard, Khorasani, and Saif}]{kordestani2023overview}
Kordestani, M., Orchard, M.~E., Khorasani, K., and Saif, M., \enquote{An overview of the state of the art in aircraft prognostic and health management strategies,} \emph{IEEE Transactions on Instrumentation and Measurement}, Vol.~72, 2023, pp. 1--15.
\newblock \doi{10.1109/TIM.2023.3236342}.

\bibitem[{Gebraeel(2006)}]{gebraeel2006sensory}
Gebraeel, N., \enquote{Sensory-updated residual life distributions for components with exponential degradation patterns,} \emph{IEEE Transactions on Automation Science and Engineering}, Vol.~3, No.~4, 2006, pp. 382--393.
\newblock \doi{10.1109/TASE.2006.876609}.

\bibitem[{Zhang et~al.(2020)Zhang, Peng, Canfei, Youren, and Zewang}]{zhang2020remaining}
Zhang, K., Peng, Z., Canfei, S., Youren, W., and Zewang, C., \enquote{Remaining useful life prediction of aircraft lithium-ion batteries based on F-distribution particle filter and kernel smoothing algorithm,} \emph{Chinese Journal of Aeronautics}, Vol.~33, No.~5, 2020, pp. 1517--1531.
\newblock \doi{10.1016/j.cja.2020.01.007}.

\bibitem[{Wang et~al.(2014)Wang, Wang, Hu, Si, and Li}]{wang2014real}
Wang, Z., Wang, W., Hu, C., Si, X., and Li, J., \enquote{A real-time prognostic method for the drift errors in the inertial navigation system by a nonlinear random-coefficient regression model,} \emph{Acta Astronautica}, Vol. 103, 2014, pp. 45--54.
\newblock \doi{10.1016/j.actaastro.2014.06.034}.

\bibitem[{Si et~al.(2012)Si, Wang, Hu, Zhou, and Pecht}]{si2012remaining}
Si, X.-S., Wang, W., Hu, C.-H., Zhou, D.-H., and Pecht, M.~G., \enquote{Remaining useful life estimation based on a nonlinear diffusion degradation process,} \emph{IEEE Transactions on reliability}, Vol.~61, No.~1, 2012, pp. 50--67.
\newblock \doi{10.1109/TR.2011.2182221}.

\bibitem[{Rodr{\'\i}guez-Pic{\'o}n et~al.(2018)Rodr{\'\i}guez-Pic{\'o}n, Rodr{\'\i}guez-Pic{\'o}n, M{\'e}ndez-Gonz{\'a}lez, Rodr{\'\i}guez-Borb{\'o}n, and Alvarado-Iniesta}]{rodriguez2018degradation}
Rodr{\'\i}guez-Pic{\'o}n, L.~A., Rodr{\'\i}guez-Pic{\'o}n, A.~P., M{\'e}ndez-Gonz{\'a}lez, L.~C., Rodr{\'\i}guez-Borb{\'o}n, M.~I., and Alvarado-Iniesta, A., \enquote{Degradation modeling based on gamma process models with random effects,} \emph{Communications in Statistics-Simulation and Computation}, Vol.~47, No.~6, 2018, pp. 1796--1810.
\newblock \doi{10.1080/03610918.2017.1324981}.

\bibitem[{Wang(2007)}]{wang2007prognosis}
Wang, W., \enquote{A prognosis model for wear prediction based on oil-based monitoring,} \emph{Journal of the Operational Research Society}, Vol.~58, No.~7, 2007, pp. 887--893.
\newblock \doi{10.1057/palgrave.jors.2602185}.

\bibitem[{Pei et~al.(2022)Pei, Si, Hu, Li, He, and Pang}]{pei2022bayesian}
Pei, H., Si, X.-S., Hu, C., Li, T., He, C., and Pang, Z., \enquote{Bayesian deep-learning-based prognostic model for equipment without label data related to lifetime,} \emph{IEEE Transactions on Systems, Man, and Cybernetics: Systems}, Vol.~53, No.~1, 2022, pp. 504--517.
\newblock \doi{10.1109/TSMC.2022.3185102}.

\bibitem[{Kumari and Wang(2024)}]{kumari2024efficient}
Kumari, L.~N., and Wang, P., \enquote{Efficient stochastic parametric estimation for lithium-ion battery performance degradation tracking and prognosis,} \emph{Journal of Manufacturing Systems}, Vol.~75, 2024, pp. 270--277.
\newblock \doi{10.1016/j.jmsy.2024.03.017}.

\bibitem[{Zhang et~al.(2021)Zhang, Yang, Li, Xiu, and Liu}]{zhang2021data}
Zhang, Y., Yang, Y., Li, H., Xiu, X., and Liu, W., \enquote{A data-driven modeling method for stochastic nonlinear degradation process with application to RUL estimation,} \emph{IEEE Transactions on Systems, Man, and Cybernetics: Systems}, Vol.~52, No.~6, 2021, pp. 3847--3858.
\newblock \doi{10.1109/TSMC.2021.3073052}.

\bibitem[{Satoh et~al.(2024)Satoh, Omata, Tsutsumi, Hashimoto, Sato, Kimura, and Abe}]{Satoh2024}
Satoh, D., Omata, N., Tsutsumi, S., Hashimoto, T., Sato, M., Kimura, T., and Abe, M., \enquote{Improved performance of data-driven simulator of liquid rocket engine under varying operating conditions,} \emph{Acta Astronautica}, Vol. 214, 2024, pp. 473--483.
\newblock \doi{10.1016/j.actaastro.2023.11.005}.

\bibitem[{Badger et~al.(2024)Badger, Barron, Cullen, and Dvorsky}]{badger2024integrated}
Badger, J., Barron, L., Cullen, P., and Dvorsky, K., \enquote{Integrated Systems and Operational Autonomy for Gateway,} \emph{International Symposium on Artificial Intelligence, Robotics and Automation in Space}, 2024.

\bibitem[{Song et~al.(2019)Song, Liu, and Zhang}]{Song2019}
Song, C., Liu, K., and Zhang, X., \enquote{A generic framework for multisensor degradation modeling based on supervised classification and failure surface,} \emph{IISE Transactions}, Vol.~51, No.~11, 2019, p. 1288–1302.
\newblock \doi{10.1080/24725854.2018.1555384}.

\bibitem[{Wen et~al.(2019)Wen, Chen, Zhao, Li, Wang, and Dou}]{Wen2019}
Wen, P., Chen, S., Zhao, S., Li, Y., Wang, Y., and Dou, Z., \enquote{A Novel Bayesian Update Method for Parameter Reconstruction of Remaining Useful Life Prognostics,} \emph{2019 IEEE International Conference on Prognostics and Health Management (ICPHM)}, 2019, p. 1–8.
\newblock \doi{10.1109/ICPHM.2019.8819377}.

\bibitem[{Gu et~al.(2021)Gu, Zhou, Zhang, Li, and Zhang}]{Gu2021}
Gu, L., Zhou, Y., Zhang, Z., Li, H., and Zhang, L., \enquote{Remaining Useful Life Prediction using Dynamic Principal Component Analysis and Deep Gated Recurrent Unit Network,} \emph{2021 Global Reliability and Prognostics and Health Management (PHM-Nanjing)}, 2021, p. 1–5.
\newblock \doi{10.1109/PHM-Nanjing52125.2021.9612895}.

\bibitem[{Kim et~al.(2019)Kim, Song, and Liu}]{Kim2019}
Kim, M., Song, C., and Liu, K., \enquote{A Generic Health Index Approach for Multisensor Degradation Modeling and Sensor Selection,} \emph{IEEE Transactions on Automation Science and Engineering}, Vol.~16, No.~3, 2019, p. 1426–1437.
\newblock \doi{10.1109/TASE.2018.2890608}.

\bibitem[{Lu et~al.(2019)Lu, Wu, Huang, and Qiu}]{lu2019aircraft}
Lu, F., Wu, J., Huang, J., and Qiu, X., \enquote{Aircraft engine degradation prognostics based on logistic regression and novel OS-ELM algorithm,} \emph{Aerospace Science and Technology}, Vol.~84, 2019, pp. 661--671.
\newblock \doi{10.1016/j.ast.2018.09.044}.

\bibitem[{Wang et~al.(2022)Wang, Liu, and Zhang}]{Wang2022}
Wang, D., Liu, K., and Zhang, X., \enquote{A Generic Indirect Deep Learning Approach for Multisensor Degradation Modeling,} \emph{IEEE Transactions on Automation Science and Engineering}, Vol.~19, 2022, p. 1924–1940.
\newblock \doi{10.1109/TASE.2021.3072363}.

\bibitem[{Che et~al.(2019)Che, Wang, Fu, and Ni}]{che2019combining}
Che, C., Wang, H., Fu, Q., and Ni, X., \enquote{Combining multiple deep learning algorithms for prognostic and health management of aircraft,} \emph{Aerospace Science and Technology}, Vol.~94, 2019, p. 105423.
\newblock \doi{10.1016/j.ast.2019.105423}.

\bibitem[{Li et~al.(2021)Li, Lei, Gebraeel, Wang, Cai, Xu, and Wang}]{Li2021}
Li, N., Lei, Y., Gebraeel, N., Wang, Z., Cai, X., Xu, P., and Wang, B., \enquote{Multi-Sensor Data-Driven Remaining Useful Life Prediction of Semi-Observable Systems,} \emph{IEEE Transactions on Industrial Electronics}, Vol.~68, No.~11, 2021, p. 11482–11491.
\newblock \doi{10.1109/TIE.2020.3038069}.

\bibitem[{Wu et~al.(2022)Wu, Zeng, Shi, Zhang, Shi, and Qin}]{Wu2022}
Wu, B., Zeng, J., Shi, H., Zhang, X., Shi, G., and Qin, Y., \enquote{Multi-sensor information fusion-based prediction of remaining useful life of nonlinear Wiener process,} \emph{Measurement Science and Technology}, Vol.~33, No.~10, 2022, p. 105106.
\newblock \doi{10.1088/1361-6501/ac7636}.

\bibitem[{Daroogheh et~al.(2017)Daroogheh, Baniamerian, Meskin, and Khorasani}]{Daroogheh2017}
Daroogheh, N., Baniamerian, A., Meskin, N., and Khorasani, K., \enquote{Prognosis and Health Monitoring of Nonlinear Systems Using a Hybrid Scheme Through Integration of PFs and Neural Networks,} \emph{IEEE Transactions on Systems, Man, and Cybernetics: Systems}, Vol.~47, No.~8, 2017, p. 1990–2004.
\newblock \doi{10.1109/TSMC.2016.2597272}.

\bibitem[{Yu(2017)}]{yu2017aircraft}
Yu, J., \enquote{Aircraft engine health prognostics based on logistic regression with penalization regularization and state-space-based degradation framework,} \emph{Aerospace Science and Technology}, Vol.~68, 2017, pp. 345--361.
\newblock \doi{10.1016/j.ast.2017.05.030}.

\bibitem[{Ibrahim et~al.(2025)Ibrahim, Eshima, Gebraeel, Nabity, and Robinson}]{ibrahim2025generative}
Ibrahim, M., Eshima, S., Gebraeel, N., Nabity, J., and Robinson, S., \enquote{A generative machine learning framework for anomaly response in cyclical processes in the ECLSS on a deep space habitat,} \emph{Acta Astronautica}, 2025.
\newblock \doi{10.1016/j.actaastro.2025.09.088}.

\bibitem[{Zhang et~al.(2014)Zhang, Hua, and Xu}]{Zhang2014}
Zhang, Q., Hua, C., and Xu, G., \enquote{A mixture Weibull proportional hazard model for mechanical system failure prediction utilising lifetime and monitoring data,} \emph{Mechanical Systems and Signal Processing}, Vol.~43, No.~1, 2014, p. 103–112.
\newblock \doi{10.1016/j.ymssp.2013.10.013}.

\bibitem[{Cox(1972)}]{Cox1972}
Cox, D.~R., \enquote{Regression Models and Life-Tables,} \emph{Journal of the Royal Statistical Society. Series B (Methodological)}, Vol.~34, No.~2, 1972, p. 187–220.

\bibitem[{Monterrubio-G{\'o}mez et~al.(2024)Monterrubio-G{\'o}mez, Constantine-Cooke, and Vallejos}]{Monterrubio2022}
Monterrubio-G{\'o}mez, K., Constantine-Cooke, N., and Vallejos, C.~A., \enquote{A review on statistical and machine learning competing risks methods,} \emph{Biometrical Journal}, Vol.~66, No.~2, 2024, p. 2300060.
\newblock \doi{10.1002/bimj.202300060}.

\bibitem[{Zhu et~al.(2016)Zhu, Yao, and Huang}]{Zhu2016}
Zhu, X., Yao, J., and Huang, J., \enquote{Deep convolutional neural network for survival analysis with pathological images,} \emph{2016 IEEE International Conference on Bioinformatics and Biomedicine (BIBM)}, 2016, p. 544–547.
\newblock \doi{10.1109/BIBM.2016.7822579}.

\bibitem[{Gupta et~al.(2019)Gupta, Sunder, Prasad, and Shroff}]{Gupta2019}
Gupta, G., Sunder, V., Prasad, R., and Shroff, G., \enquote{CRESA: A Deep Learning Approach to Competing Risks, Recurrent Event Survival Analysis,} \emph{Advances in Knowledge Discovery and Data Mining}, edited by Q.~Yang, Z.-H. Zhou, Z.~Gong, M.-L. Zhang, and S.-J. Huang, Springer International Publishing, Cham, 2019, p. 108–122.
\newblock \doi{10.1007/978-3-030-16145-3_9}.

\bibitem[{Marthin and Tutkun(2023)}]{Marthin2023}
Marthin, P., and Tutkun, N.~A., \enquote{Recurrent neural network for complex survival problems,} \emph{Journal of Statistical Computation and Simulation}, Vol.~0, No.~0, 2023, p. 1–25.
\newblock \doi{10.1080/00949655.2023.2176504}.

\bibitem[{Rollock and Klaus(2025)}]{rollock2025characterizing}
Rollock, A.~E., and Klaus, D.~M., \enquote{Characterizing the impact of emergent technologies on earth communications reliance for crewed deep space missions,} \emph{Acta Astronautica}, Vol. 226, 2025, pp. 803--813.
\newblock \doi{10.1016/j.actaastro.2024.11.011}.

\bibitem[{Ulusoy and Reisman(2025)}]{ulusoy2025investment}
Ulusoy, U., and Reisman, G., \enquote{Investment construct in human autonomy teaming for deep space habitat operations,} \emph{Acta Astronautica}, Vol. 236, 2025, pp. 117--127.
\newblock \doi{10.1016/j.actaastro.2025.06.037}.

\bibitem[{Aremu et~al.(2020)Aremu, Cody, Hyland-Wood, and McAree}]{Aremu2020}
Aremu, O.~O., Cody, R.~A., Hyland-Wood, D., and McAree, P.~R., \enquote{A relative entropy based feature selection framework for asset data in predictive maintenance,} \emph{Computers \& Industrial Engineering}, Vol. 145, 2020, p. 106536.
\newblock \doi{10.1016/j.cie.2020.106536}.

\bibitem[{Kim and Liu(2021)}]{Kim2021}
Kim, M., and Liu, K., \enquote{A Bayesian deep learning framework for interval estimation of remaining useful life in complex systems by incorporating general degradation characteristics,} \emph{IISE Transactions}, Vol.~53, No.~3, 2021, p. 326–340.
\newblock \doi{10.1080/24725854.2020.1766729}.

\bibitem[{Jiao et~al.(2020)Jiao, Peng, Dong, and Zhang}]{Jiao2020}
Jiao, R., Peng, K., Dong, J., and Zhang, C., \enquote{Fault monitoring and remaining useful life prediction framework for multiple fault modes in prognostics,} \emph{Reliability Engineering \& System Safety}, Vol. 203, 2020, p. 107028.
\newblock \doi{10.1016/j.ress.2020.107028}.

\bibitem[{Chehade et~al.(2018)Chehade, Song, Liu, Saxena, and Zhang}]{Chehade2018}
Chehade, A., Song, C., Liu, K., Saxena, A., and Zhang, X., \enquote{A data-level fusion approach for degradation modeling and prognostic analysis under multiple failure modes,} \emph{Journal of Quality Technology}, Vol.~50, No.~2, 2018, p. 150–165.
\newblock \doi{10.1080/00224065.2018.1436829}.

\bibitem[{Wu and Li(2023)}]{Wu2023}
Wu, H., and Li, Y.-F., \enquote{A Multi-Sensor Fusion-based Prognostic Model for Systems with Partially Observable Failure Modes,} \emph{IISE Transactions}, Vol.~0, No.~ja, 2023, p. 1–21.
\newblock \doi{10.1080/24725854.2023.2222402}.

\bibitem[{Fu et~al.(2025)Fu, Kwon~Huh, and Liu}]{Fu2025}
Fu, Y., Kwon~Huh, Y., and Liu, K., \enquote{Degradation Modeling and Prognostic Analysis Under Unknown Failure Modes,} \emph{IEEE Transactions on Automation Science and Engineering}, Vol.~22, 2025, pp. 11012--11025.
\newblock \doi{10.1109/TASE.2025.3530845}.

\bibitem[{Su and Fang(2025)}]{su2025deep}
Su, Y., and Fang, X., \enquote{Deep Learning-Based Residual Useful Lifetime Prediction for Assets with Uncertain Failure Modes,} \emph{Journal of Computing and Information Science in Engineering}, 2025, pp. 1--29.
\newblock \doi{10.1115/1.4067843}.

\bibitem[{Zhang et~al.(2026)Zhang, Manikkan, Krishnan, Azimi, Vaccino, Wang, Xue, Dyke, Bilionis, Han, and Parsons}]{zhang2025managing}
Zhang, Z., Manikkan, S., Krishnan, M., Azimi, M., Vaccino, L., Wang, J., Xue, C., Dyke, S.~J., Bilionis, I., Han, S., and Parsons, P., \enquote{Managing delay-induced challenges in remote monitoring of uncrewed space habitats: The impact of forecasting telemetry visualizations,} \emph{Acta Astronautica}, Vol. 241, 2026, pp. 1--18.
\newblock \doi{10.1016/j.actaastro.2025.12.036}.

\bibitem[{Jiang and Wang(2010)}]{Jiang2010}
Jiang, C.-R., and Wang, J.-L., \enquote{Covariate Adjusted Functional Principal Components Analysis for Longitudinal Data,} \emph{The Annals of Statistics}, Vol.~38, No.~2, 2010, p. 1194–1226.
\newblock \doi{10.1214/09-AOS742}.

\bibitem[{Karhunen(1947)}]{Karhunen1947}
Karhunen, K., \emph{Über lineare Methoden in der Wahrscheinlichkeitsrechnung}, Kirjapaino oy. sana, 1947.
\newblock Google-Books-ID: bGUUAQAAIAAJ.

\bibitem[{Yao et~al.(2005)Yao, Müller, and Wang}]{Yao2005}
Yao, F., Müller, H.-G., and Wang, J.-L., \enquote{Functional Data Analysis for Sparse Longitudinal Data,} \emph{Journal of the American Statistical Association}, Vol. 100, No. 470, 2005, p. 577–590.
\newblock \doi{10.1198/016214504000001745}.

\bibitem[{Städler et~al.(2010)Städler, Bühlmann, and van~de Geer}]{Stadler2010}
Städler, N., Bühlmann, P., and van~de Geer, S., \enquote{L1-Penalization for Mixture Regression Models,} \emph{TEST}, Vol.~19, No.~2, 2010, p. 209–256.
\newblock \doi{10.1007/s11749-010-0197-z}.

\bibitem[{Cleveland(1979)}]{Cleveland2012}
Cleveland, W.~S., \enquote{Robust Locally Weighted Regression and Smoothing Scatterplots,} \emph{Journal of the American Statistical Association}, 1979.
\newblock \doi{10.1080/01621459.1979.10481038}.

\bibitem[{M{\"u}ller and Zhang(2005)}]{muller2005time}
M{\"u}ller, H.-G., and Zhang, Y., \enquote{Time-varying functional regression for predicting remaining lifetime distributions from longitudinal trajectories,} \emph{Biometrics}, Vol.~61, No.~4, 2005, pp. 1064--1075.
\newblock \doi{10.1111/j.1541-0420.2005.00378.x}.

\bibitem[{Fang et~al.(2015)Fang, Zhou, and Gebraeel}]{Fang2015}
Fang, X., Zhou, R., and Gebraeel, N., \enquote{An adaptive functional regression-based prognostic model for applications with missing data,} \emph{Reliability Engineering \& System Safety}, Vol. 133, 2015, p. 266–274.
\newblock \doi{10.1016/j.ress.2014.08.013}.

\bibitem[{Frederick et~al.(2007)Frederick, DeCastro, and Litt}]{frederick2007user}
Frederick, D.~K., DeCastro, J.~A., and Litt, J.~S., \enquote{User's guide for the commercial modular aero-propulsion system simulation (C-MAPSS),} Tech. rep., 2007.

\bibitem[{Saxena et~al.(2008)Saxena, Goebel, Simon, and Eklund}]{Saxena2008}
Saxena, A., Goebel, K., Simon, D., and Eklund, N., \enquote{Damage propagation modeling for aircraft engine run-to-failure simulation,} \emph{2008 International Conference on Prognostics and Health Management}, 2008, p. 1–9.
\newblock \doi{10.1109/PHM.2008.4711414}.

\bibitem[{Murphy(2012)}]{murphy2012}
Murphy, K.~P., \emph{Machine learning: a probabilistic perspective}, MIT press, 2012.

\end{thebibliography}
